\documentclass[10pt,twocolumn,letterpaper]{article}

\usepackage[pagenumbers]{cvpr}

\usepackage[numbers,sort&compress]{natbib}

\usepackage{tabularx}
\newcolumntype{C}{>{\centering\arraybackslash}X}
\usepackage{bbm}

\usepackage{enumitem}
\setlist[itemize]{align=right,itemindent=1em,labelsep=2pt,labelwidth=1em,leftmargin=0pt,nosep}

\usepackage{multirow}
\usepackage{colortbl}

\colorlet{colorFst}{Green!25}      
\colorlet{colorSnd}{SpringGreen!45}
\colorlet{colorTrd}{Yellow!30}    
\colorlet{colorLow}{darkgray!30} 
\newcommand{\fs}{\cellcolor{colorFst}\bf}
\newcommand{\nd}{\cellcolor{colorSnd}}
\newcommand{\rd}{\cellcolor{colorTrd}}

\newcommand{\boldparagraph}[1]{\vspace{2pt}\noindent{\bf #1}}

\newcommand{\figref}[1]{Fig.~\ref{#1}}
\newcommand{\eqnref}[1]{Eq.~\eqref{#1}}
\newcommand{\tabref}[1]{Tab.~\ref{#1}}
\newcommand{\secref}[1]{Sec.~\ref{#1}}

\newif\ifcommented
\commentedtrue

\ifcommented
    \newcommand{\martin}[1]{\noindent\textcolor{ForestGreen}{\textbf{[Martin:~}#1\textbf{]}}}
    \newcommand{\jie}[1]{\noindent\textcolor{Dandelion}{\textbf{[Jie:~}#1\textbf{]}}}
    \newcommand{\manuel}[1]{\noindent\textcolor{Purple}{\textbf{[Manuel:~}#1\textbf{]}}}
    \newcommand{\lixin}[1]{\noindent\textcolor{OliveGreen}{\textbf{[Lixin:~}#1\textbf{]}}}    
\else
    \newcommand{\martin}[1]{\unskip}
    \newcommand{\jie}[1]{\unskip}
    \newcommand{\manuel}[1]{\unskip}
    \newcommand{\lixin}[1]{\unskip}
\fi

\newcommand{\methodname}{ODHSR\xspace}

\definecolor{cvprblue}{rgb}{0.21,0.49,0.74}
\usepackage[pagebackref,breaklinks,colorlinks,allcolors=cvprblue]{hyperref}

\def\paperID{16431}
\def\confName{CVPR}
\def\confYear{2025}

\title{\methodname: Online Dense 3D Reconstruction of Humans and Scenes\\ from Monocular Videos}

\author{Zetong Zhang\textsuperscript{1} \quad Manuel Kaufmann\textsuperscript{1} \quad Lixin Xue\textsuperscript{1} \quad Jie Song\textsuperscript{1,2,3} \quad Martin R. Oswald\textsuperscript{4} \\
    \\
    \textsuperscript{1}ETH Zürich \quad 
    \textsuperscript{2}HKUST(GZ) \quad
    \textsuperscript{3}HKUST \quad
    \textsuperscript{4}University of Amsterdam\\
}

\begin{document}

\twocolumn[{
  \maketitle
  \vspace{-1.5em}
  \centering
  \captionsetup{type=figure}
  \includegraphics[width=0.95
  \textwidth]{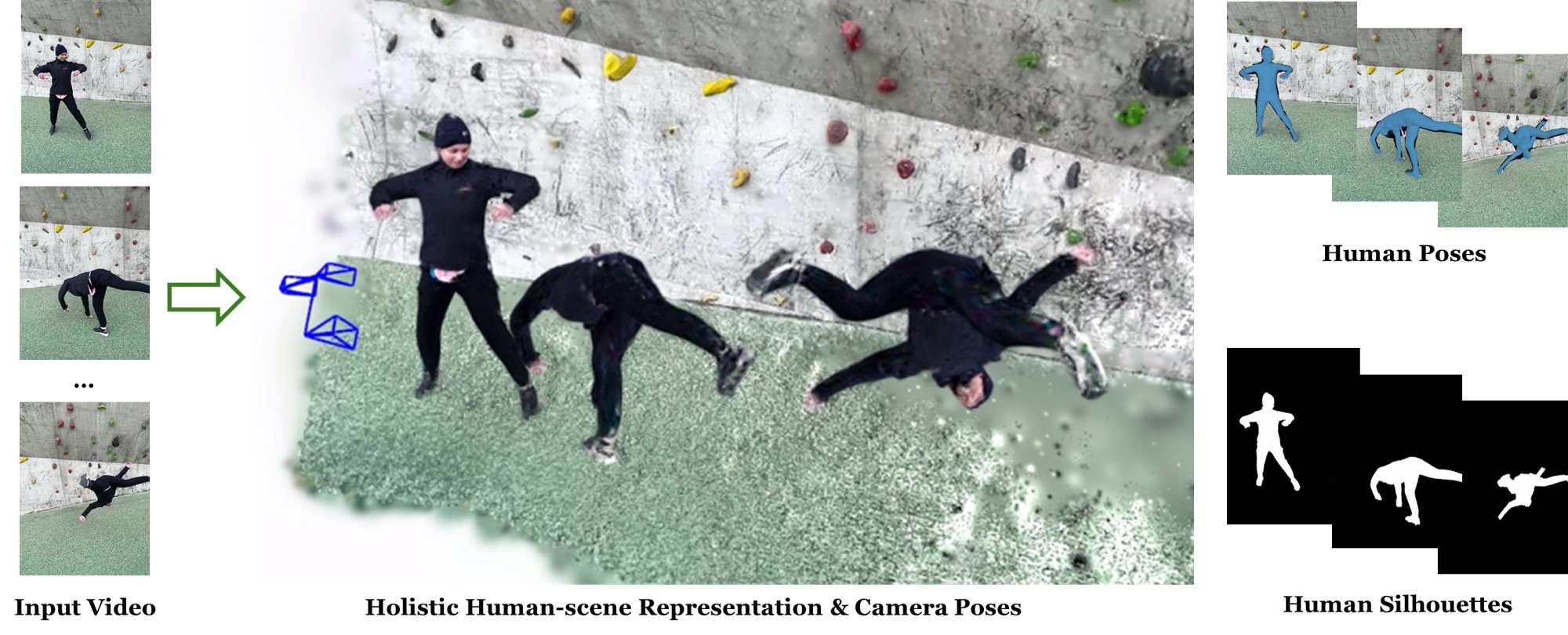}\\[-4pt]
  \captionof{figure}{\textbf{\methodname} takes monocular RGB input videos of humans and jointly reconstructs a photorealistic dense Gaussian representation of the scene and the moving human as well as camera poses, human poses, and human silhouettes within a SLAM setting.}
  \label{fig:teaser}
  \vspace{2em}
}]

\begin{abstract}
Creating a photorealistic scene and human reconstruction from a single monocular in-the-wild video figures prominently in the perception of a human-centric 3D world. Recent neural rendering advances have enabled holistic human-scene reconstruction but require pre-calibrated camera and human poses, and days of training time. In this work, we introduce a novel unified framework that simultaneously performs camera tracking, human pose estimation and human-scene reconstruction in an online fashion. 3D Gaussian Splatting is utilized to learn Gaussian primitives for humans and scenes efficiently, and reconstruction-based camera tracking and human pose estimation modules are designed to enable holistic understanding and effective disentanglement of pose and appearance. Specifically, we design a human deformation module to reconstruct the details and enhance generalizability to out-of-distribution poses faithfully. Aiming to learn the spatial correlation between human and scene accurately, we introduce occlusion-aware human silhouette rendering and monocular geometric priors, which further improve reconstruction quality. Experiments on the EMDB and NeuMan datasets demonstrate superior or on-par performance with existing methods in camera tracking, human pose estimation, novel view synthesis and runtime. Our project page is at \href{https://eth-ait.github.io/ODHSR}{https://eth-ait.github.io/ODHSR}.
\end{abstract}    

\section{Introduction}
Building robotic agents for supporting humans in every-day tasks requires the efficient and holistic understanding of scenes and dynamic humans in an online manner. Previous works in this domain do not always meet these criteria: They either focus on the human reconstruction~\cite{guo2023vid2avatar,qian20233dgsavatar} or the scene reconstruction alone \cite{teed2022droidslamdeepvisualslam,Zhu2023NICER,sandstrom2024splat,Matsuki2024MonoGS}, or if combined, they are rather slow to compute (order of days) \cite{xue2024hsr} or require pre-calibrated cameras~\cite{hugs}.

We present the first unified framework that only needs a monocular RGB video to simultaneously perform camera tracking, human pose estimation and dense photorealistic reconstruction of both the scene and the human. Our system, called \methodname, is orders of magnitude (75x) faster than previous work \cite{xue2024hsr} and constitutes an online system.

To achieve this, we carefully design a 3DGS-based~\cite{kerbl3Dgaussians} optimization that \emph{jointly} tracks camera poses and human poses as well as a consistent dense scene map with both geometry and appearance information. We model the human avatar with 3D Gaussians in canonical space, guided by SMPL-based deformations~\cite{SMPL:2015}, whereby the deformation is decomposed into rigid and non-rigid parts to account for dynamic garments.
The avatar model is designed to be simple and effective for tracking of the human pose. The robust and accurate recovery of both scene geometry and human shape in combination with the decomposition of human motion and camera motion from only RGB input is highly challenging.
To this end, we design a Gaussian Splatting-based SLAM pipeline based on \cite{Matsuki2024MonoGS} with effective camera tracking and mapping capabilities due to the incorporation of a monocular geometric prior~\cite{depth_anything_v2} as well as human shape and pose priors from SMPL estimates to initialize the relative positioning between the human and the scene. Specifically, we introduce an occlusion-aware human silhouette term to aid the decomposition of human and scene reconstruction. To keep the pipeline efficient, only a small number of keyframes is maintained for instant mapping. The 3DGS-based formulation over implicit functions allows for direct gradient flow to the Gaussians and facilitates better generalization to unseen poses.

To assess our method, we compare \methodname to several baselines on two datasets: Neuman~\cite{jiang2022neuman}, and EMDB~\cite{kaufmann2023emdb} which feature challenging in-the-wild sequences with reference SMPL and camera poses.
\methodname outperforms the state of the art in terms of reconstruction quality for both the human avatar and scene reconstruction. It does so without access to any ground-truth depth, with an online algorithm, and while achieving superior runtime and real-time rendering. Further, the experiments show that \methodname outperforms the state of the art in global human pose estimation and achieves on-par performance in camera tracking.

In summary, we contribute the first method that simultaneously performs dense and detailed reconstruction of both human avatars and the 3D scene in an online manner from only a monocular RGB video. This is enabled by leveraging the direct gradient flow of a holistic 3DGS-based parametrization and an effective joint optimization. The proposed methodology is robust, accurate and efficient in comparison to existing works.

\section{Related Work}

\boldparagraph{3D Scene Reconstruction and SLAM.}
Modern scene reconstruction methods span grid-based, point-based, network-based, and hybrid approaches. Grid-based methods~\cite{sun2021neucon, Weder_2020_CVPR, mueller2022instant} are memory-intensive and rely on predefined resolutions, while point-based methods \cite{orbeez-slam, bad-slam} adapt dynamically to surfaces, reducing memory waste but facing connectivity issues. Network-based approaches~\cite{mildenhall2020nerf, yariv2021volume} provide detailed scene representations but struggle with scalability and efficiency, especially in online tasks. Hybrid techniques~\cite{mueller2022instant} aim to balance speed and quality. Simultaneous Localization and Mapping (SLAM) reconstructs scenes while tracking camera trajectories. Dense SLAM methods are either frame-centric or map-centric. Frame-centric approaches \cite{Czarnowski:2020:10.1109/lra.2020.2965415, teed2022droidslamdeepvisualslam} minimize photometric errors for frame-to-frame motion but lack global consistency. Map-centric methods build unified 3D representations, utilizing global information for tracking and reconstruction. Classical techniques use voxel grids \cite{kinectfusion}, point clouds \cite{Sandström2023ICCV,zhang2024glorie}, or neural feature grids~\cite{Zhu2022CVPR, Zhu2023NICER} for 3D representation.

Recently, 3D Gaussian splatting with differentiable rendering capabilities has emerged as an effective scene representation~\cite{kerbl3Dgaussians}. Its tile-based renderer enables faster training and rendering than NeRF-based methods~\cite{mildenhall2020nerf}, making it well-suited for dense real-time Gaussian-based SLAM~\cite{keetha2024splatam,yugay2023gaussianslam, Matsuki2024MonoGS,yan2024gs,huang2024photo}, see \cite{tosi2024nerfs} for an overview. Splat-SLAM~\cite{sandström2024splatslamgloballyoptimizedrgbonly} augments RGB-only SLAM~\cite{Matsuki2024MonoGS} with a monocular prior, a deformable Gaussian map to incorporate dense bundle adjustment and loop closures for more accurate tracking. We built our method upon the RGB-only approach~\cite{Matsuki2024MonoGS} and extend it with a monocular prior as well as human pose and shape priors for effective joint reconstruction and tracking.

\boldparagraph{Human Pose Estimation (HPE).}
With the emergence of differentiable statistical body models like SMPL~\cite{SMPL:2015}, and powerful Deep Learning methods, substantial progress has been achieved in recent years. Landmark papers include SMPLify~\cite{bogo2016smplautomaticestimation3d}, which proposes to optimize the 3D poses to 2D keypoint detections, and HMR~\cite{hmrKanazawa17}, which directly regresses 3D poses from an image using 2D keypoint supervision and adversarial losses.
Many other works have followed in these footsteps, achieving impressive results \cite{lassner2017unitepeopleclosingloop,humanMotionKZFM19,pavlakos2018learningestimate3dhuman,omran2018neuralbodyfittingunifying,kocabas2019vibe, pavlakos2019smplx, pymaf2021,song2020lgd,Sun2021ROMP,Li2021hybrik,Cho2022FastMETRO,li2022cliff,lin2021-mesh-graphormer,mehta2017vnect,zhou2016sparseness,Dwivedi2024TokenHMR,stathopoulos2024scoreHMR,goel2023humans,Kocabas2021PARE,joo2020eft}.
Traditionally, these works focus on camera-relative pose estimation.
More recently, it was proposed to disentangle the camera from the human motion and estimate the human pose in global coordinates \cite{yuan2022glamr,ye2023slahmr,TRACE,shin2023wham,shen2024gvhmr,kocabas2024pace}.
This is similar to our setting as we also estimate human pose and camera motion in global space. However, we additionally reconstruct a dense scene and a photo-realistic appearance of the human. To deal with depth-ambiguities and occlusions in images, others have proposed to use body-worn sensors for motion capture, \eg, \cite{vonMarcard2018,PIPCVPR2022,Trumble:BMVC:2017,kaufmann2021empose}. This increases instrumentation requirements, but the collected poses can serve as reference data. We evaluate our method on the EMDB~\cite{kaufmann2023emdb} in-the-wild dataset.

\boldparagraph{3D Human Reconstruction.}
The above mentioned works all estimate the naked human body or assume tight clothing.
It has also been proposed to reconstruct the clothing and appearance, \eg, by extending explicit SMPL-based representations \cite{alldieck2019tex2shape,bhatnagar2019mgn,Xiang2020MonoClothCap,xiu2023econ}, by using a personalized pre-scanned template of the human \cite{MonoPerfCap_SIGGRAPH2018,habermann2020deepcap,yu2019SimulCap}, articulated NeRFs \cite{jiang2022neuman,weng_humannerf_2022_cvpr}, or implicit surface fields \cite{jiang2022selfrecon,guo2023vid2avatar,saito2021scanimate,xiu2022icon,saito2020PIFuHD}. 
All of these works either neglect the scene, reconstruct low-quality geometry, or impose large computational costs. With the overwhelming success of 3D Gaussian Splatting~\cite{kerbl3Dgaussians}, the community quickly adopted the representation for human avatar modeling, both from multi-view \cite{zielonka2023drivable3dgaussianavatars, li2024animatablegaussians} and monocular images \cite{liu2023animatable,hugs,hu2023gauhuman,shao2024splattingavatar,qian20233dgsavatar} to reduce computation times. Animatable 3D Gaussian~\cite{liu2023animatable} uses multi-resolution hash grids to predict the Gaussian attributes and achieves fast training, but lacks robust to input pose errors because it does not optimize for the poses. \cite{shao2024splattingavatar} introduces hybrid mesh and Gaussian representation, which leads to a similar problem because there are no direct gradients to the input pose parameters.
\cite{hugs} and \cite{hu2023gauhuman} learn LBS weights to map from canonical to the posed space, regularized by SMPL.  However, they do not model any local deformations that might be due to non-tight clothing.
3DGS-Avatar~\cite{qian20233dgsavatar} explicitly models the local non-rigid deformation and pose-dependent color change with two MLPs, but the large MLP architecture slows down its training compared to other Gaussian-based works.

\boldparagraph{Human-Scene Modeling.}
Modeling humans and scenes together is a topic studied in differing shades.
Some methods do so from an egocentric perspective with the use of body-worn sensors and cameras \cite{Guzov2021HPS,EgoLocate2023,Zhang2022EgoBody,Dai2022HSC4D,Dai2023Slopr4d} but they either require a pre-scanned scene or do not reconstruct the detailed human avatar.
Other work focuses on exocentric views, and solves tasks such es extracting interaction graphs \cite{savva2016pigraphs}, disambiguating human poses with scene constraints \cite{Hassan2019PROX,Li_3DV2022}, placing humans into existing scenes~\cite{zhang2020place}, or modeling rich contacts using static multi-view setups~\cite{Huang2022RICH}.
Closest to our work are HSR~\cite{xue2024hsr} and HUGS~\cite{hugs} who operate in the same setting. HSR extends the success of neural implicit shape functions to jointly model the human and the scene. This formulation incurs high computational costs.
Like ours, HUGS models the human avatar and the scene with 3D Gaussians. However, its triplane formulation is ill-suited for an online setting as converging to good features for the tracking is slow.

\section{Method}
\begin{figure*}[!t]
    \centering
     \includegraphics[width=0.95\textwidth]{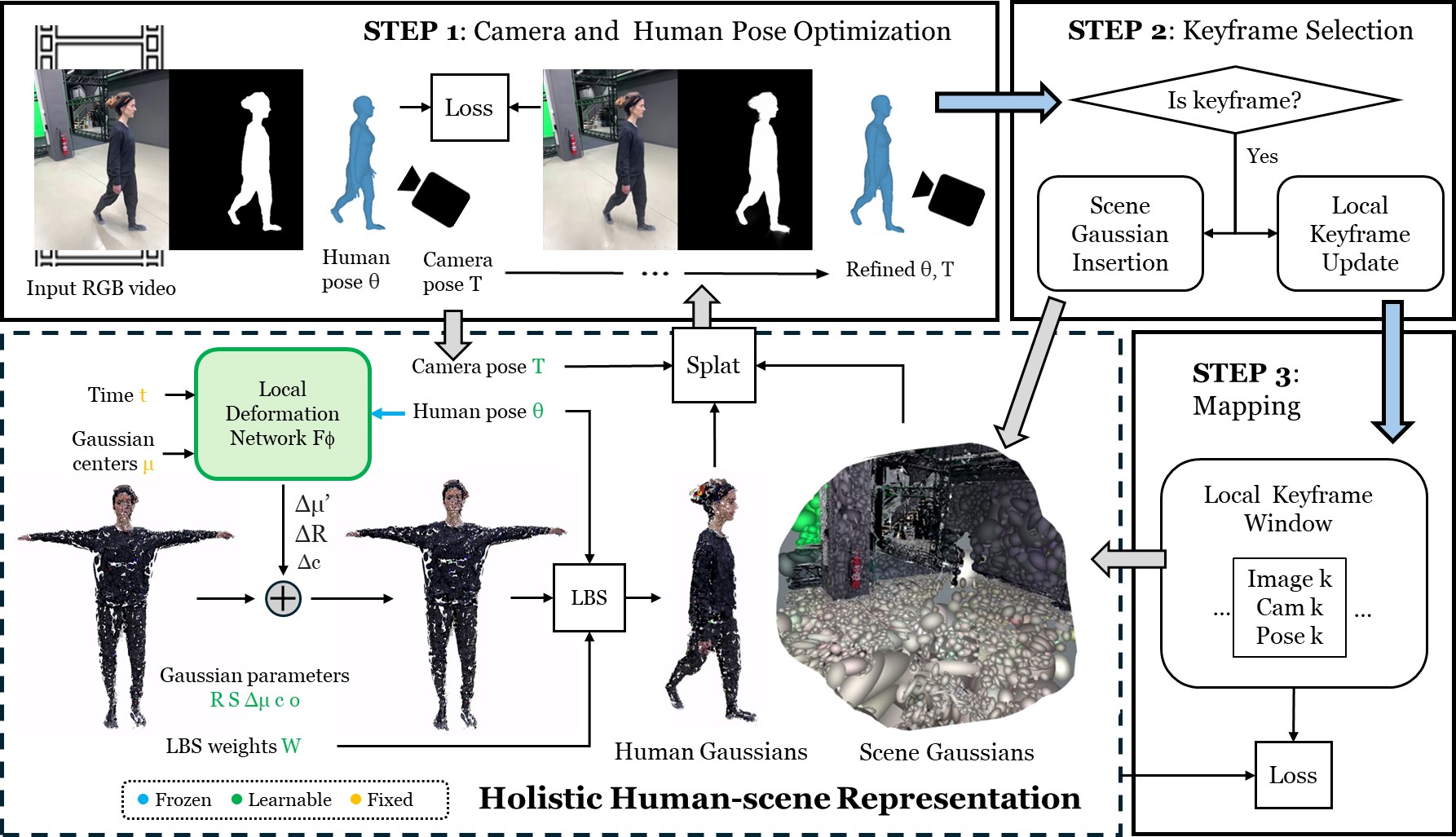}  
     \caption{\textbf{System Overview of \methodname.} Given a monocular video featuring a human in the scene, we simultaneously track the camera and human poses for each frame while training 3D Gaussian primitives. Camera and human pose optimization is achieved through dense matching for view synthesis and leveraging monocular geometric cues. Mapping is carried out within a small local keyframe window, and we apply multiple regularizations to enhance reconstruction quality from the sparse set of keyframes.} \label{fig:overview}
\end{figure*}

We first describe how we model the human avatar in \secref{sec:method_avatar_rep}, and how this ties in with the chosen scene representation in \secref{sec:method_scene_rep}. Finally, in \secref{sec:method_slam} we describe how we simultaneously perform camera tracking, human pose estimation and human-scene reconstruction.
For an overview of our method, please refer to \figref{fig:overview}.

\subsection{3D Avatar Representation} \label{sec:method_avatar_rep}
We represent the human body in the canonical space with a set of 3D Gaussians $G_H = \{G_{H,i} | i = 1,\ldots, N_H\}$, where each Gaussian $G_{H,i}$ is parameterized by its own center position $\mu_{H}$, center offset $\Delta\mu_{H}$, rotation $R_{H}$, scale $S_{H}$, opacity $o_{H}$, RGB color $c_{H}$ and Linear Blend Skinning (LBS) weights $W_{H} \in \mathbb{R}^{J}$ with respect to $J$ SMPL joints. Among these parameters, centers $\mu_{H}$ are initially sampled around SMPL vertices and stay fixed at the initial positions, and we instead optimize the center offsets$\Delta\mu_{H}$, along with all the other parameters. The skeletal deformation and skinning driven by SMPL can only model the rigid deformation of human joints, but dynamic garments may not precisely follow the joint deformations. Therefore, similar to \cite{qian20233dgsavatar, weng_humannerf_2022_cvpr, moreau2024humangaussiansplattingrealtime}, we decompose the Gaussian deformation into a rigid part and a non-rigid part.

\boldparagraph{Time-pose Dependent Non-rigid Deformation and Appearance.}
The deformations of the irregular garment and hair are dependent on the human pose and time-accumulating dynamics. Hence, we model per-Gaussian local deformation via a time-pose conditioned multi-resolution hash encoding network \cite{mueller2022instant}. Moreover, shadows cast on the human surface change with the geometric deformation. A second multiresolution hash encoding network is introduced to model the ambient occlusion factor to address the shadow issue. We denote the network parameters as $F_{\phi}$ and refer the readers to the supplementary material for the detailed input encoding and architecture design. Given the current time step $t$ and human pose $\theta \in \mathbb{R}^{3J}$, we obtain the local deformation $\Delta\mu'_{H}$, $\Delta R_{H}$ and ambient occlusion $\Delta {c}$ of each Gaussian via:
\begin{equation}
    \Delta\mu'_{H}, \Delta R_{H}, \Delta c_{H} = F_\phi(\mu{_H}, t, \theta)
\end{equation}
The canonical Gaussians are then deformed by:
\begin{align}
    \mu_{H,d} &= \mu_{H} +\Delta\mu_{H} + \Delta\mu'_{H}  \label{eq:center_d}\\
    R_{H,d}  &= R_H \cdot \Delta R_H \label{eq:rotation_d} \\
     c_{H,d} &= \Delta c_{H} \cdot c_{H}
\end{align}

\boldparagraph{Rigid Transformation.}
Following SMPL, we use LBS to deform the model using the joints in the underlying SMPL model defined by shape $\beta \in \mathbb{R}^{10}$ and pose parameters $\theta \in \mathbb{R}^{3J}$. The transformation $M_j \in SE(3)$ of each joint $j$ from the canonical space to the posed space is calculated using the kinematic tree. Each Gaussian's skinning transformation $P$ is the weighted average of joint transformations according to learnable parameter $W_H$, formulated as $P = \sum_{j=1}^{J} W_{H,j}M_j$, where $W_{H,j}$ is the $j$-th element of LBS weight $W_{H} \in \mathbb{R}^{J}$ corresponding to the $j$-th joint. We then transform the canonical Gaussian positions and rotations calculated in Eq.~\eqref{eq:center_d}, \eqref{eq:rotation_d} to the world frame as follows:
\begin{align}
    \mu_{H,w} &= P \mu_{H,d} \\
    R_{H,w}   &= P_{:3,:3} R_{H,d}
\end{align}

\subsection{Holistic Human-scene Representation} \label{sec:method_scene_rep}
\boldparagraph{Holistic Representation.}
We use standard 3D Gaussians to model the scene. The set of scene Gaussians is denoted as $G_S = \{G_{S,i} |  i = 1,\ldots, N_S\}$, where $G_{S,i}$ is the $i$-th scene Gaussian and is composed of its own center $\mu_{S}$, scale $S_{S}$, rotation $R_{S}$, opacity $o_{S}$ and RGB color $c_{S}$.

Given scene Gaussians $G_S$ and transformed human Gaussians in the world frame $G_H$, we merge them into a global Gaussian set $G = G_S + G_H$ as the holistic human-scene primitives and feed them into the Gaussian rasterizer to render the color map $\hat{I}$, depth map $\hat{D}$, and opacity map $\hat{O}$, respectively.

\boldparagraph{Occlusion-aware Human Silhouette.}
3D Gaussians can, by design, handle the occlusion between objects since the Gaussians are sorted by depth along the camera ray in the rasterizer. The occlusion-aware human opacity (silhouette) $\hat{O}_H$ can then be retrieved as:
\begin{equation}
    \hat{O}_H = \sum_{j=1}^{N_H} \alpha_j \prod_{k=1}^{N_j}(1-\alpha_k) \label{eq:sil}
\end{equation}
where $N_j$ is the number of all the human and scene Gaussians whose depth along this pixel ray is smaller than that of $G_{H,j}$ and $\prod_{k=1}^{N_j}(1-\alpha_k)$ represents for the transmittance at $j$-th human Gaussian calculated from all the Gaussians in front. By taking into account the scene Gaussians, which are closer to the camera, we obtain the human silhouette rendering where the occlusion is faithfully modeled.

\subsection{SLAM} \label{sec:method_slam}
In this section, we present the details of our full SLAM framework, where camera tracking, human pose estimation, and dense human-scene reconstruction are performed simultaneously.
For each frame, we compute the residuals between the input and the rendering from the holistic Gaussian representation to track both the camera and human pose. A keyframe check is performed on each tracked frame, and a local keyframe window is updated, with which we run mapping to jointly reconstruct the human and the scene. In the end, we follow the idea of adopting global bundle adjustment in SLAM approaches\cite{teed2022droidslamdeepvisualslam, sandstrom2024splat, yugay2023gaussianslam, Matsuki2024MonoGS} to finetune the holistic representation with all the keyframes. Our pipeline contains two threads for efficiency. The tracking thread takes in new frames, runs camera and human pose optimization and selects keyframes, while the mapping thread simultaneously runs mapping and bundle adjustment over the local keyframe window.

\boldparagraph{Initialization.}
The estimation of high-dimensional human pose and shape parameters $\theta$, $\beta$ of a person from an RGB image is challenging without prior knowledge. We start with the poses from an off-the-shelf monocular human pose estimator, WHAM \cite{shin2023wham}. For the very first frame of the sequence, we carefully refine the human pose estimate $\theta$ in a model-free approach by minimizing the 2D keypoint loss. Following \cite{kaufmann2023emdb}, we extract $N = 25$ 2D keypoints from ViTPose \cite{xu2022vitpose} denoted by $\Tilde{x}_i$ and define a 2D keypoint loss as
\begin{equation}
    L_\text{kp} = \sum_{i=1}^{25} \mathbbm{1}[\text{conf}_{i} > 0.5]\cdot\rho(\hat{x}_i - \Tilde{x}_i) \label{eq:keypoint}
\end{equation}
where $\hat{x_i} = K[R|t]X_i$ are projection of SMPL joints $X_i$ onto the image and $\text{conf}_{i}$ is the confidence of the $i$-th keypoint predicted by the keypoint detector, $\mathbbm{1}$ is the indicator function and $\rho$ is the robust Geman-McClure function \cite{Geman1987StatisticalMF}.

In order to faithfully recover the spatial correlation between the human and the scene and to produce trustworthy scene geometry for subsequent camera and human pose estimation, we utilize a monocular depth estimator \cite{depth_anything_v2} to obtain per-frame monocular depth estimation $\Tilde{D}$. We solve for the scale and shift parameters $w,b$ for the first frame disparity $1/\Tilde{D}$ by aligning it with the SMPL mesh disparity calculated from $\beta$ and $\theta$ with a RANSAC estimator to reduce the effect of outliers. We then initialize the scene Gaussians at the positions inferred from the re-scaled monocular depth of the first frame, i.e., $1/(w/\Tilde{D} + b)$.

\boldparagraph{Camera and Human Pose Estimation.}
Given each new image $I$, we jointly optimize the camera pose $T$ and human pose $\theta$ of the current frame via the following optimization constraints while keeping the holistic representation fixed. 
\begin{itemize}
    \item \textbf{RGB Loss.} We minimize the photometric residual between the input $I$ and rendered image $\hat{I}$ as follows. 
    \begin{equation}
        L_\text{rgb} = {\lVert I -\hat{I} \rVert}_1 \label{eq:loss_rgb}
    \end{equation}
    
    \item \textbf{Optical Flow Loss.}
    Following \cite{Zhu2023NICER, sandstrom2024splat}, to avoid the local minima introduced by the pixel-wise RGB loss, we estimate the optical flow $\Tilde{p}_{ij}$ from the last keyframe $i$ and the current frame $j$ with a pretrained estimator \cite{xu2022gmflow}. Given the rendered depth $\hat{D}$ of frame $i$, we can also compute the flow from pixels $p_{i}$ in frame $i$ to projected pixel coordinates in frame $j$ and minimize the optical flow loss $L_\text{flow}$. 
    \begin{equation}
        L_\text{flow} = {\lVert \Tilde{p}_{ij} - K\Delta T_{ij} \hat{D}_{i}K^{-1}{[p_{i},1]}^{\top} \rVert }_1
    \end{equation}
    where $K$ is the camera intrinsic matrix, $\Delta T_{ij}$ is the relative pose between frames $i,j$. Since this consistency only holds for static objects, we mask out the dynamic human via a pre-estimated human segmentation mask from \cite{kirillov2023segment}. We keep the pose of keyframe $i$ fixed and expect this loss to only contribute to the camera pose optimization.  
    
    \item \textbf{Monocular Depth Loss.} Inspired by \cite{xue2024hsr}, we make use of geometric priors from pre-trained depth estimators and enforce the depth consistency between our rendered depth $\hat{D}$ and the monocular depth $\Tilde{D}$. Since the monocular depth is usually prone to error in far objects, like sky and buildings, in the outdoor scene. During tracking, to stabilize the pose optimization, we compute inverse depth map $d = 1 / D$ and minimize the geometric residual between the rendering and monocular ones as:
    \begin{equation}
        L_\text{disp} = {\lVert \hat{d} - (w\Tilde{d}+ b)\rVert }_1 \label{eq:disp}
    \end{equation}
    where $w, b \in \mathbb{R}$ are the scale and shift used to align $\hat{d}$ and $\Tilde{d}$, since $\Tilde{d}$ is only known up to an unknown scale. We solve for $w$ and $b$ per image with least squares at each optimization iteration, where both the human and scene pixels are utilized to ensure the scaled monocular depth map can faithfully reflect the human-scene spatial correlations.
    \item \textbf{Human Silhouette Loss.} The noisy color and depth rendering from the online mapping could make the optimization converge slowly. We also utilize a human silhouette loss as an auxiliary signal for the human pose optimization. Given the pre-estimated human segmentation $\Tilde{O}_H$ and rendered human silhouette $\hat{O}_H$ as in \eqnref{eq:sil}, we formulate the human silhouette loss as follows.
    \begin{equation}
        L_\text{sil} = {\lVert \hat{O}_H - \Tilde{O}_H \rVert}_1 \label{eq:loss_sil}
    \end{equation}
    \item \textbf{2D Keypoint Loss.} RGB loss suffers from color ambiguities due to the sparse texture on the human, and thus is not sufficient to accurately align the human joints and learn the human poses. We additionally use the 2D keypoint loss formulated in \eqnref{eq:keypoint} for each frame as an auxiliary term to guide the joint alignment.
\end{itemize} 
The final loss for the joint camera and human pose optimization is the weighted sum of all the losses introduced above:
\begin{align}
    \label{eq:loss_pose}
    L_\text{pose} = &\lambda_\text{rgb} L_\text{rgb} + \lambda_\text{flow} L_\text{flow} \\ \nonumber
    & + \lambda_\text{disp} L_\text{disp} +\lambda_\text{sil} L_\text{sil} +  \lambda_\text{kp} L_\text{kp}
\end{align}
Notably, for $L_\text{rgb}$, $L_\text{flow}$ , $L_\text{disp}$ and $L_\text{sil}$, we use whole-image rendered opacity map $\hat{O}$ as the pixel weights and compute weighted $l1$ loss to mitigate the effect from unseen regions.

\boldparagraph{Keyframing.}
After tracking, each frame will be checked for keyframe registration based on multiple criteria, including frame interval, camera displacement, human joint displacements, and Gaussian co-visibility. These criteria are designed to find the most informative frames for the mapping. Refer to the supplementary material for the details. 

Following the strategy of Gaussian Splatting SLAM~\cite{Matsuki2024MonoGS}, we only maintain a small number of keyframes in the current window $\mathcal{W}_k$ and update the window constantly to only keep frames that are either the latest or have the largest visual overlap. By doing this, we update the Gaussians and networks with the knowledge from the new keyframes, which can be generalized better to a subsequent frame. 

For each new keyframe, we insert new Gaussians into the scene by back-projecting the re-scaled monocular depth of the static background to 3D space to capture newly visible scene components. 

\boldparagraph{Mapping.}
During mapping, keyframes in local window $\mathcal{W}_k$ along with two random past keyframes are used to reconstruct recently visible regions and avoid forgetting the global map. To enforce the consistency between observation and reconstruction, we minimize the following re-rendering losses:
\begin{itemize}
    \item \textbf{RGB Loss.} We use $l1$ RGB loss for color reconstruction.
    \item \textbf{Human Silhouette Loss.} Separating the human from the scene is challenging, especially when the views are limited in our online mapping setting. We also use the human silhouette loss formulated in \eqnref{eq:loss_sil} for reconstruction.
    \item \textbf{Monocular Depth Loss.} We use monocular depth loss $L_\text{depth}$ to stabilize and clean the scene to prevent ``floaters'' appearing in free space, which could occlude the human in the camera view. Different from \eqnref{eq:disp}, we keep $w,b$ fixed, compute the absolute depth residual, and only constrain the depth rendering of scene pixels during mapping.
\end{itemize}

Due to the limited training poses in the local keyframe window, we introduce several regularizations on the avatar representation to better generalize novel human poses and reconstruct animatable avatars. 
\begin{itemize}
    \item \textbf{Local Deformation Loss.} We penalize the magnitude of the local deformation and ambient occlusion factor to stabilize the training with  $L_\text{deform}$.
    \item \textbf{LBS Weights Loss.} To prevent the skinning weights $W_H$ from overfitting on the training poses, we supervise the per-Gaussian skinning weight with the skinning weight in the SMPL model via $L_\text{LBS}$.
    \item \textbf{Canonical Center Loss.} We softly regularize the geometry of the reconstructed human with the underlying SMPL model. With this regularization, $L_\text{center}$, we prevent the Gaussians from moving too wildly due to limited training views and poses.
\end{itemize}

To summarize, we minimize the weighted sum of all these losses in the mapping thread:
\begin{align}
  \label{eq:loss_map}
  L_\text{map} = &\lambda_\text{rgb}L_\text{rgb} + \lambda_\text{sil}L_\text{sil} + \lambda_\text{depth}L_\text{depth}  \\  \nonumber
  &+ \lambda_\text{LBS}L_\text{LBS}
  + \lambda_\text{center}L_\text{center} + \lambda_\text{deform}L_\text{deform} 
\end{align}

Furthermore, the hash encoding network-based deformation and appearance modules exhibit sensibility to noise when exposed to novel pose and time encodings. To mitigate this issue, we propose two training strategies aimed at effectively learning network parameters with robust interpolation and extrapolation properties across the spatial and temporal dimensions, thereby stabilizing the mapping process.
Please refer to the supplementary material for further details on these loss and strategy designs.

\section{Experiments}

\subsection{Experimental Setup}

\boldparagraph{Datasets.} The following datasets are used for evaluation:
\begin{itemize}
    \item \textbf{EMDB dataset}~\cite{kaufmann2023emdb} is a recently published large-scale in-the-wild dataset consisting of versatile sequences captured in outdoor or indoor scenes. We identify five distinct sequences that presented various challenges, such as extended human and camera trajectories, human occlusions, prominent shadows on the human body, sparse background texture within the lab setting, and unconventional human poses (\eg cartwheels). For consistency concerns, we take the first 500 frames, \ie, the first 16 seconds, from all these sequences in our experiments. This is our major dataset for quantitative evaluation. 
    
    \item \textbf{NeuMan dataset}~\cite{jiang2022neuman} is an in-the-wild dataset with six sequences, each captured with a moving camera that pans through the scene. Our keyframe selection is deactivated for this dataset, and we instead follow the dataset split outlined in \cite{jiang2022neuman} and only run evaluation for reconstruction quality.
\end{itemize}

\boldparagraph{Metrics.}
We report standard photometric rendering quality metrics (PSNR, SSIM, LPIPS) on the non-keyframes/test set for the novel view synthesis task. These metrics are calculated on both whole images and human-only images. Moreover, we care about the accuracy of our predicted poses. For camera tracking, we follow conventional monocular SLAM evaluation protocol to align trajectory and calculate trajectory error (ATE RMSE). The predicted human poses are evaluated as well via local pose metrics MPJPE, PA-MPJPE and MVE and global motion metrics W-MPJPE and WA-MPJPE. Following EMDB~\cite{kaufmann2023emdb}, we report a jitter metric to take account of the smoothness of the estimated joint trajectories. 

\begin{table*}[t]
    \centering
    \footnotesize
    \begin{tabularx}{\textwidth}{lCCCCCCCC}
    \toprule
    & \multicolumn{3}{c}{\textbf{Whole images}} & \multicolumn{3}{c}{\textbf{Human-only regions}} & \multicolumn{2}{c}{\textbf{FPS}} \\
    & PSNR$\uparrow$ & SSIM$\uparrow$ & LPIPS$\downarrow$& PSNR$\uparrow$ & SSIM$\uparrow$ & LPIPS$\downarrow$ & Training$\uparrow$ & Rendering$\uparrow$ \\
    \midrule
    GauHuman~\cite{hu2023gauhuman} & - & - & - & 25.313&0.943 & 0.057 & \nd 0.150 & \fs 150\\
    3DGS-Avatar~\cite{qian20233dgsavatar} & - &- & - & 27.952 & \fs 0.967 &0.035 & 0.112 & \rd 60\\
    Vid2Avatar~\cite{guo2023vid2avatar} & 16.656 & 0.413 & 0.599 & 24.258 & 0.948 & 0.061 &  $<10^{-3}$ & 0.02 \\
    HUGS~\cite{hugs} & \rd 21.605 & \rd 0.659 & \fs 0.181 & \rd 26.165 & 0.947 & \rd 0.033 & 0.042  & 40\\
    HSR~\cite{xue2024hsr} & 18.675& 0.463&0.632&25.127&0.924&0.054& 0.002 & 0.05 \\ \midrule
    Ours (DROID-SLAM) & \nd 23.458 & \nd 0.756 & \rd 0.200 & \fs 29.203  & \rd 0.965  & \fs 0.030 & \fs 0.181 & \nd 85 \\
    Ours (Full) & \fs 23.790  & \fs 0.767  & \nd 0.197   & \nd 28.955 & \nd 0.966 &  \nd 0.031 & \rd 0.141 & \nd 85 \\ 

    \bottomrule
    \end{tabularx}
    \caption{\textbf{Novel view synthesis evaluation on the EMDB dataset~\cite{kaufmann2023emdb}.} For the human-only setting, we render the avatar on white background for all the baselines and compute metrics over the whole image. We report training and rendering FPS, with training FPS calculated as the inverse of the average training time per image. All experiments were run on a single Nvidia GeForce RTX 4090 GPU.}
    \label{tab:emdb_psnr_holistic}
\end{table*}

\boldparagraph{Reconstruction Baselines.}
Since, to the best of our knowledge, our work is the first online dense human-scene reconstruction method in the community, we opt for offline reconstruction works for baseline comparison. Our auto-selected keyframes are used for training, while the remaining frames are used for evaluation only. We compare our method with the holistic human-scene reconstruction method HSR~\cite{xue2024hsr}, Vid2Avatar~\cite{guo2023vid2avatar} and HUGS~\cite{hugs}. For 3DGS-based~\cite{kerbl3Dgaussians} scene reconstruction approach HUGS, we additionally provide captured depth maps to initialize the scene Gaussians. Our online approach contrarily omits such demanding preprocessing. In addition, we also compare our method with 3DGS-based human-only reconstruction works 3DGS-Avatar~\cite{qian20233dgsavatar} and GauHuman~\cite{hu2023gauhuman}. These offline works typically require known camera poses. For fair comparison on the EMDB dataset, we use the SotA tracker DROID-SLAM~\cite{teed2022droidslamdeepvisualslam} with these baselines, as well as ours while disabling the camera tracking module. This evaluation strategy also provides insights into the performance of our camera tracker. On the NeuMan dataset, we evaluate the baselines directly using ground truth camera poses. Consistent with our approach, all baselines take WHAM estimates as the human pose initialization.

\subsection{Evaluation Results}

\begin{table*}[t]
    \centering
    \footnotesize
    \begin{tabularx}{\textwidth}{lCCCCCCCC}
    \toprule
    & \multicolumn{3}{c}{\textbf{Whole images}} & \multicolumn{3}{c}{\textbf{Human-only regions}} & \multicolumn{2}{c}{\textbf{FPS}} \\
    & PSNR$\uparrow$ & SSIM$\uparrow$ & LPIPS$\downarrow$& PSNR$\uparrow$ & SSIM$\uparrow$ & LPIPS$\downarrow$ & Training$\uparrow$ &Rendering$\uparrow$\\
    \midrule
    GauHuman~\cite{hu2023gauhuman}  & - & - & - & 30.731 & 0.977 & \rd 0.017 & \rd 0.038 & \fs 150 \\
    3DGS-Avatar~\cite{qian20233dgsavatar} & - & - & - & \fs 32.920 & \fs 0.988 & \fs 0.015 & 0.015 & \rd 60 \\
    Vid2Avatar~\cite{guo2023vid2avatar} & 15.640 & 0.551 & 0.572 & 30.967 & \nd 0.981 & 0.018 & $<10^{-3}$ & 0.008 \\
    HUGS~\cite{hugs} & \nd 26.667 & \nd 0.851 & \fs0.126  &30.136 & 0.977& 0.017 & 0.019 & 40\\
    HSR~\cite{xue2024hsr} & 21.676 & 0.669 & 0.526 &29.033&0.971&0.026& $<10^{-3}$ & 0.01 \\ \midrule
    Ours (GT camera) & \fs 27.784 & \fs 0.870 & \nd0.153  &\nd 32.079 & \nd 0.981 & \nd 0.016 & \fs  0.046 & \nd 85 \\
    Ours (Full) & \rd 26.470 & \rd 0.825  & \rd 0.174   &  \rd 31.729   & \nd 0.981 & \rd  0.017 & \nd 0.040 & \nd 85 \\
    \bottomrule
    \end{tabularx}
    \caption{\textbf{Novel view synthesis evaluation on the NeuMan dataset~\cite{jiang2022neuman}.} Compared to recent offline methods, our approach provides better whole image rendering performance while being comparable on human-only regions despite our more challenging online setting.}
    \label{tab:neuman_psnr_all}
\end{table*}

\begin{figure*}[!htb]
    \centering
    \begin{subfigure}[b]{0.13\textwidth}
    \centering
        \includegraphics[width=\textwidth]{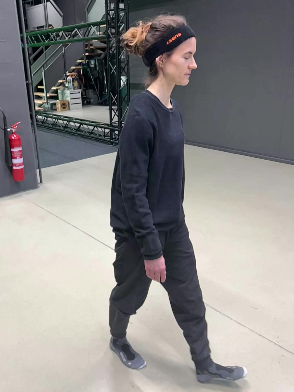}
    \end{subfigure}
    \begin{subfigure}[b]{0.13\textwidth}
    \centering
        \includegraphics[width=\textwidth]{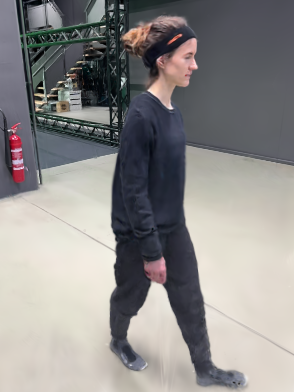}  
    \end{subfigure}
    \begin{subfigure}[b]{0.13\textwidth}
    \centering
        \includegraphics[width=\textwidth]{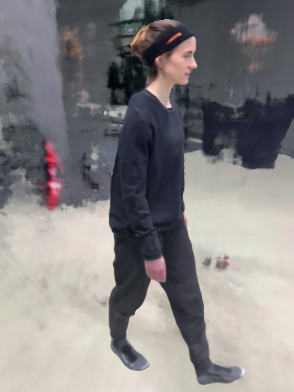} 
    \end{subfigure}
    \begin{subfigure}[b]{0.13\textwidth}
    \centering
        \includegraphics[width=\textwidth]{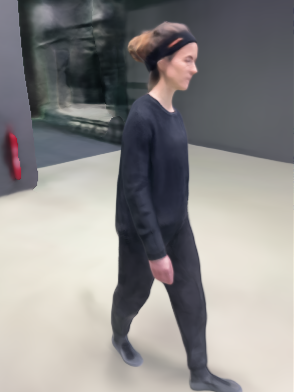}    
    \end{subfigure}
    \begin{subfigure}[b]{0.13\textwidth}
    \centering
        \includegraphics[width=\textwidth]{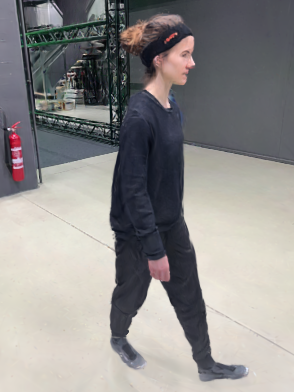}    
    \end{subfigure}
     \begin{subfigure}[b]{0.13\textwidth}
    \centering
        \includegraphics[width=\textwidth]{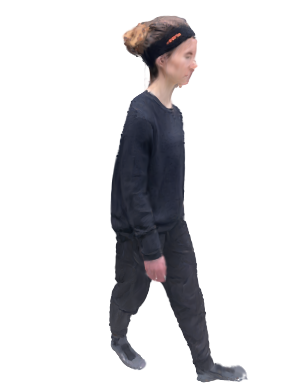} 
    \end{subfigure}
    \begin{subfigure}[b]{0.13\textwidth}
    \centering
        \includegraphics[width=\textwidth]{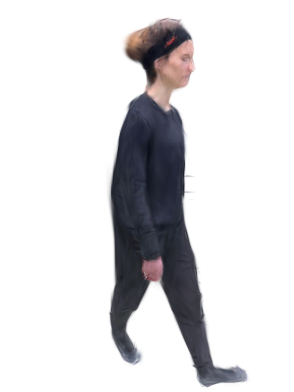} 
    \end{subfigure}
    \vskip\baselineskip

    \begin{subfigure}[b]{0.13\textwidth}
    \centering
        \includegraphics[width=\textwidth]{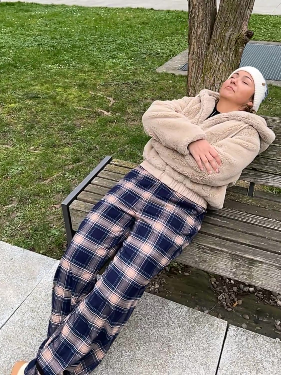} 
        \caption[]{GT}
    \end{subfigure}
    \begin{subfigure}[b]{0.13\textwidth}
    \centering
        \includegraphics[width=\textwidth]{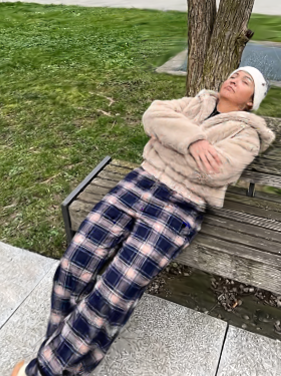} 
         \caption[]{Ours}
    \end{subfigure}
    \begin{subfigure}[b]{0.13\textwidth}
    \centering
        \includegraphics[width=\textwidth]{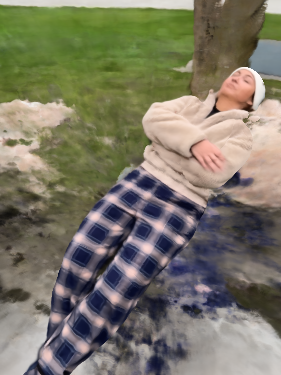}  
        \caption[]{Vid2Avatar}
    \end{subfigure}
    \begin{subfigure}[b]{0.13\textwidth}
    \centering
        \includegraphics[width=\textwidth]{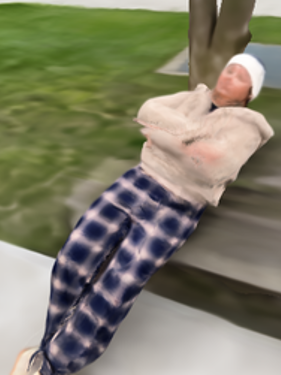} 
        \caption[]{HSR}
    \end{subfigure}
    \begin{subfigure}[b]{0.13\textwidth}
    \centering
        \includegraphics[width=\textwidth]{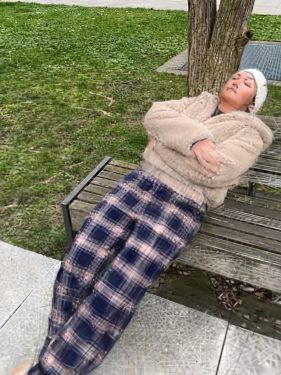} 
        \caption[]{HUGS}
    \end{subfigure}
     \begin{subfigure}[b]{0.13\textwidth}
    \centering
        \includegraphics[width=\textwidth]{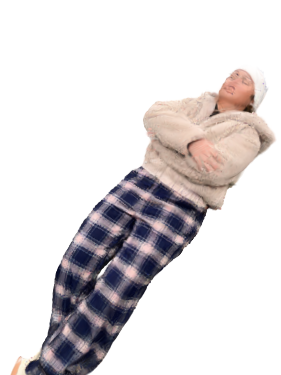}
         \caption[]{3DGS-Avatar}
    \end{subfigure}
    \begin{subfigure}[b]{0.13\textwidth}
    \centering
        \includegraphics[width=\textwidth]{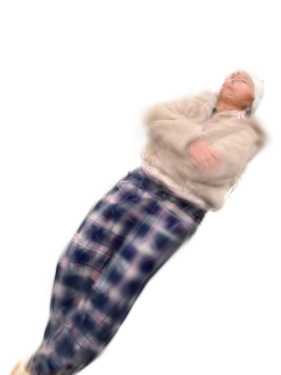} 
         \caption[]{GauHuman}
    \end{subfigure}
    \caption{\textbf{Qualitative results on the EMDB dataset~\cite{kaufmann2023emdb}.} Our online approach is highly competitive when compared to recent offline methods and outperforms most of them especially with respect to image sharpness and data fidelity.}
    \label{fig:novel_view_emdb}
\end{figure*}

\boldparagraph{Novel View Synthesis.}
In \tabref{tab:emdb_psnr_holistic} and \tabref{tab:neuman_psnr_all}, we compare our reconstruction quality over the whole image and human-only renderings on the EMDB and NeuMan dataset. We present two variants of our approach, either running the camera tracking with the optimization scheme or using an existing SOTA camera tracker as an alternative. We observe that our camera tracking module achieves comparable or even better reconstruction while operating in a fully online manner.
\methodname overperforms all the holistic human-scene reconstruction methods by a large margin. The scene model in Vid2Avatar is designed to be human-centric, and the scene geometry and textures are not properly learned from multi-view correlations. HSR extends Vid2Avatar with scene field and holistic representation but still shows degraded performance for the view synthesis task in outdoor scenes. HUGS achieves moderate performance in the background, but their overall results are worse than ours as their human reconstruction is prone to pose noises. We additionally compare the human-only reconstruction quality with two 3DGS-based methods. On the challenging EMDB dataset where the input poses are noisy and there exists drastic illumination change and garment deformations, we show significant advantages over the baselines. On the NeuMan dataset, we notice a noticeable drop in performance when switching from known camera poses. Even without the known camera trajectory, our holistic reconstruction achieves the highest performance, and our avatar reconstruction quality is second only to 3DGS-Avatar while requiring significantly less time.

We further show the qualitative results in \figref{fig:novel_view_emdb}. In general, our method shows the best reconstruction quality with complete contours and vivid photorealistic textures, where the detailed clothes' deformation and shadows are better recovered than others. Vid2Avatar struggles with bad background reconstruction and missing body parts, and produces a lot of artifacts. HSR handles the background modeling better, but its SDF representation is prone to oversmooth features. HUGS fails to produce decent reconstructions that faithfully recover the detailed human appearance because their joint Gaussian and human pose optimization schemes can not effectively disentangle the appearance and pose. 3DGS-Avatar performs well in approximating clothes wrinkles and reserving color smoothness but faces the same problem as HUGS. Gauhuman uses an MLP to learn the pose correction offset during training and performs the worst at handling input pose inconsistency. The renderings become quite blurry, and the details of the clothes and faces are almost lost. 

\boldparagraph{Camera Tracking.} To provide a straightforward insight into our camera tracker, we evaluate predicted camera trajectories against the ground truth on the EMDB dataset. Our method achieves an ATE of 8.4 cm, which is comparable to the SotA DROID-SLAM, while owns a unique advantage in scale estimation by leveraging humans as a reference. We further show that incorporating human information explicitly into the tracker facilitates ours to overperforms static SLAM approaches where humans are masked out.

\boldparagraph{Human Pose Estimation.}
We demonstrate that our framework enhances the accuracy of the human trajectory. Using WHAM's local (camera) frame estimation for pose initialization, we evaluate our refined human trajectory against their predicted global trajectories and compare it with raw WHAM predictions. Our method achieves a WA-MPJPE of 175.215 mm on the EMDB dataset, while WHAM records 636.001 mm. 

For detailed evaluation of camera tracking and human pose estimation, please refer to the supplementary material.

\subsection{Ablation Study}

\begin{table}[ht!]
  \centering
  \footnotesize
  \setlength{\tabcolsep}{6 pt}
  \begin{tabular}{lccccc}
    \toprule
    &  PSNR$\uparrow$ & ATE RMSE [m]$\downarrow$ & WA-MPJPE [mm]$\downarrow$\\
    \midrule
    w/o $L_\text{flow}$     & 22.593 & 0.214  & 301.621 \\
    w/o $L_\text{keypoint}$ &  22.263  & \nd 0.121 & \nd 230.875 \\
    w/o $L_\text{disp}$     & \nd 22.769 &  0.165  & 252.547 \\
    w/o $L_\text{sil}$      & \rd 22.648 & \rd  0.148 &  \rd 240.838 \\
    Full model              & \fs 23.790 & \fs 0.084 & \fs 175.215 \\
    \bottomrule
  \end{tabular}
  \caption{\textbf{Ablation study of camera and human pose optimization on the EMDB dataset~\cite{kaufmann2023emdb}.} The view synthesis, camera tracking and human pose estimation results demonstrate consistently superior performance for the full model.}
  \label{tab:abl_track}
\end{table}

\begin{figure}[!htb]
    \centering
    \begin{subfigure}[b]{0.15\textwidth}
    \centering
        \includegraphics[width=\textwidth]{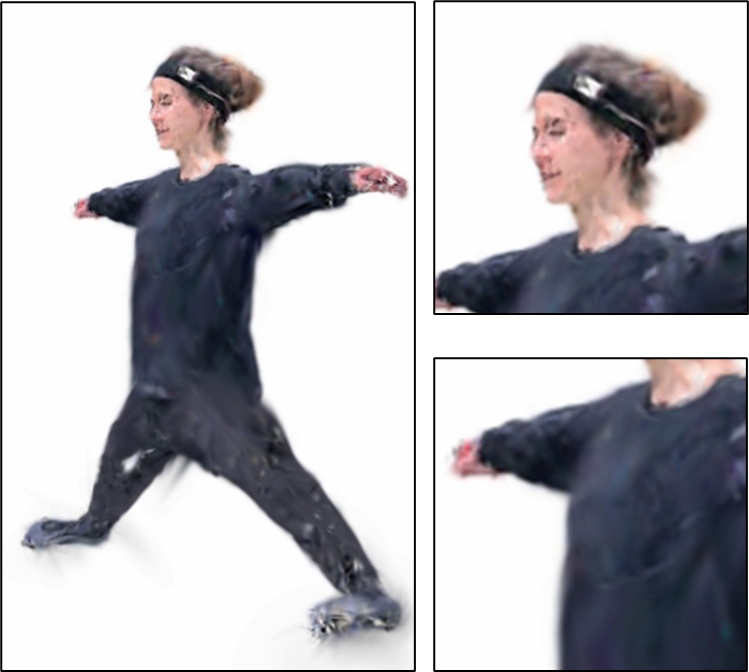} 
         \caption[]{Full model}      
    \end{subfigure}
    \begin{subfigure}[b]{0.15\textwidth}
        \centering 
        \includegraphics[width=\textwidth]{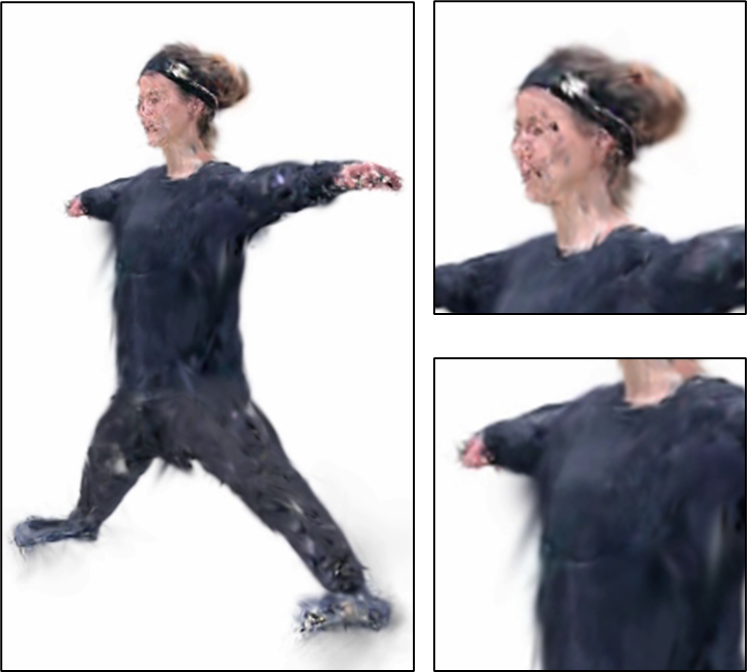} 
        \caption[]{w/o $L_{\text{center}}$}
    \end{subfigure}
    \begin{subfigure}[b]{0.15\textwidth}
        \centering 
        \includegraphics[width=\textwidth]{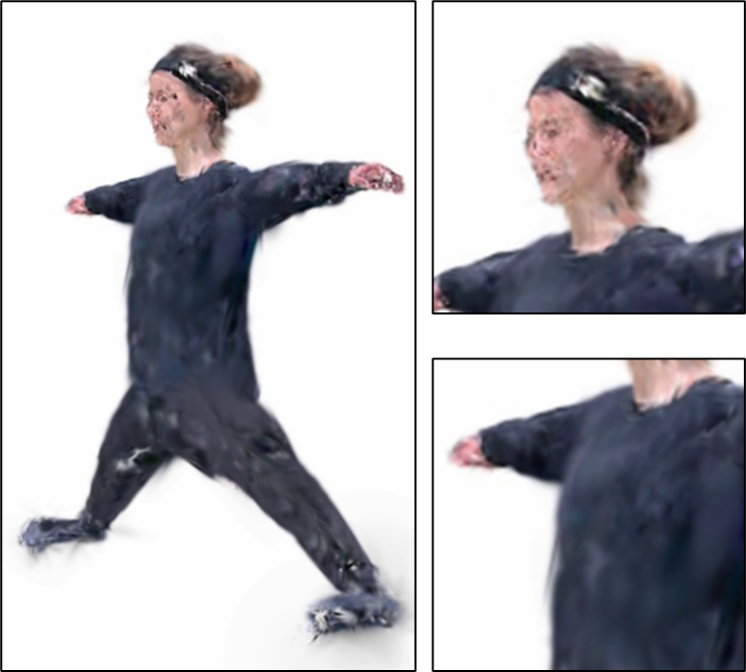} 
        \caption[]{w/o $L_{\text{LBS}}$}   
    \end{subfigure}
    \caption{\textbf{Qualitative ablation of regularizations on avatar}. Our full model comprises the least amount of artifacts.} 
    \label{fig:abl_avatar_reg}
\end{figure}

We conduct ablation experiments to examine the effect of removing each of the loss components in \eqnref{eq:loss_pose}. The 2D keypoint loss and human silhouette loss work together to accelerate human pose optimization, which subsequently facilitates monocular depth alignment. Clean scene geometry, achieved through precise depth scaling, further reduces errors in optical flow, monocular depth, and silhouette accuracy. Together, these four interconnected losses are fundamental to the success of our proposed approach. 

Alongside tracking, we inspect the impact of our designed objective, particularly its effect on human subjects, which enables effective generalization to new perspectives and poses even with a limited training dataset. As illustrated in \figref{fig:abl_avatar_reg}, omitting the canonical center loss $L_{\text{center}}$ and LBS loss $L_{\text{LBS}}$ allows Gaussian deformation to go unrestrained, significantly compromising reconstruction accuracy, especially in challenging areas like the face and arms. 
\section{Conclusion}

We introduce \methodname, the first unified framework capable of simultaneously performing camera localization, human pose estimation, and dense human-scene reconstruction from monocular RGB videos in a fully online setting. By integrating monocular geometric priors with explicit 3D primitives, our approach effectively models human-scene spatial correlations, enhancing pose optimization and reconstruction accuracy. Our joint optimization demonstrates improved performance in novel view synthesis and human pose estimation tasks, marking a significant step forward for real-time, monocular video-based 3D reconstruction.

\newpage
\section*{Acknowledgments}
This work was partially supported by the Swiss SERI Consolidation Grant "AI-PERCEIVE". Computations were performed on the ETH Zürich Euler Cluster. We thank Chen Guo and TianJian Jiang for their valuable suggestions in this research project.

{
    \small
    \bibliographystyle{ieeenat_fullname}
    \bibliography{main}
}

\clearpage
\newpage

\renewcommand{\thesection}{\Alph{section}}
\setcounter{table}{3}
\setcounter{figure}{4}
\setcounter{equation}{14}
\setcounter{section}{0}

\section{Time-pose Dependent Deformation Network} \label{sec:deform_nn}

\begin{figure}[!h]
    \centering
     \includegraphics[width=0.5\textwidth]{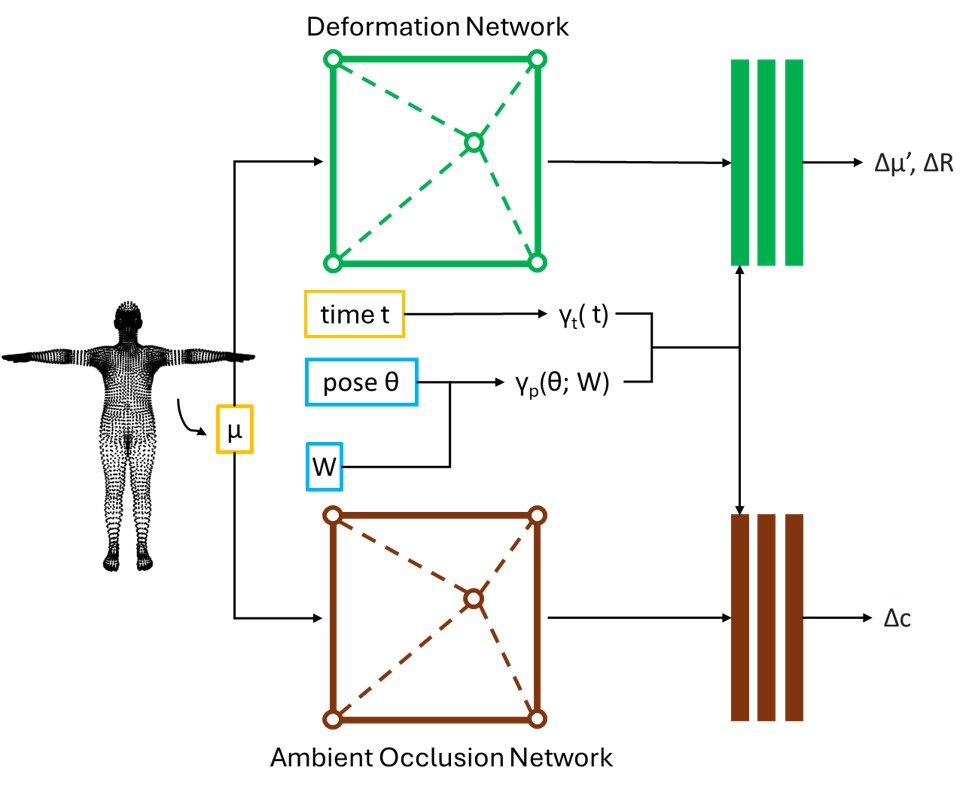}  
     \caption{Local deformation and ambient occlusion network. \textcolor{yellow!85!black}{Yellow}: Fixed parameters; \textcolor{cyan}{Blue}: Frozen parameters.} \label{fig:deform_nn}
\end{figure}

In this section, we present the details of our designed time-pose dependent non-rigid deformation and appearance module introduced in \secref{sec:method_avatar_rep} in the main paper. \footnote{In \secref{sec:deform_nn} and \secref{sec:losses}, we omit the subscripts of human Gaussian parameters for simplicity.}

Our network design is shown in Fig.\ref{fig:deform_nn}, where two parallel multiresolution hash encoding networks are utilized to learn geometric and photometric deformation respectively. Given time step $t$, human pose parameter $\theta$ and per-Gaussian LBS weight $W$, the encodings of time and pose are denoted as $\gamma_t(t)$ and $\gamma_p(\theta;W)$, respectively. Specifically, we use the positional encoding as in ~\cite{mildenhall2020nerf} to encode the normalized input time, where the max degree is set to be 4. For the pose, we follow the idea of ~\cite{moreau2024humangaussiansplattingrealtime} to use an attention-weighting scheme to encode only the pose parameters of joints that are close to the Gaussian center. By doing this, the redundant information in the global pose parameters $\theta$ can be removed so that the local deformation around the input Gaussian will be better learned without the spurious correlation of irrelevant joints. This is inspired by SCANimate~\cite{Saito:CVPR:2021}, which uses the LBS weight and a predefined attention map $V$ that limits the propagation of deformation within four neighboring joints in the kinematic tree. The pose encoding is then formulated as follows:
\begin{equation}
    \gamma_p(\theta;W) = (V \cdot W) \odot \theta
\end{equation}

where $\theta$ is in quaternion format and $\odot$ denotes element-wise multiplication.

Similar to ~\cite{liu2023animatable}, we use the fixed Gaussian centers $\mu$ as the input of the hash grid to compress the size of the hash table and prevent optimization from diverging owing to the unstable Gaussian displacements. The time and pose encodings are concatenated with the hash encoding features queried with the Gaussian center $\mu$ as the input of the shallow MLP networks to produce deformation $\Delta\mu', \Delta R$ and 1-channel ambient occlusion $\Delta c$ prediction. With the MLP architecture, we use its smoothness prior and expect good interpolation properties to be learned to facilitate generalization to novel frames and poses. In practice, we parameterize the rotation $ \Delta R$ in the form of a quaternion vector and limit the ambient occlusion factor within the range of 0-2.

\section{Keyframe Management}
In this section, we introduce our designed criteria for keyframe selection selection.

\begin{itemize}
    \item \textbf{Frame interval}: Only a frame whose time difference from the last keyframe is above a threshold $\tau_t$ can be chosen as a new keyframe so that we can avoid registering the keyframes too frequently and idling the main thread for a long period.
    \item \textbf{Camera motion}: If the displacement of the current camera from that of the last keyframe is larger than a threshold $\tau_c$, we will add the current frame to the keyframe set to span a wide baseline.
    \item \textbf{Human motion}: We measure the averaged human joint displacement from the last keyframe for each frame to estimate the pose change. Large human motion is likely to lead to unobserved local non-rigid deformation and appearance change. Thus, we register frames with drastic pose change, where the joint displacement is above $\tau_j$, to better model the garment deformation.
    \item \textbf{Gaussian covisibility}: 3D Gaussians respect visibility ordering since the rasterizer will sort Gaussians along the camera ray. Similar to ~\cite{Matsuki2024MonoGS}, we mark Gaussians as visible if they contribute to the rendering from the camera view. We then compute the covisibility of all the Gaussians (human + scene) by computing the IOU value of visible ones between the current frame and the last keyframe. If the covisibility is below a threshold $\tau_v$, the current frame will be selected as the new keyframe to reduce redundant visual overlap between keyframes. 

When a new keyframe is added or the size of the local keyframe window is larger than $\tau_s$, we update the window with the new keyframe.
Previous keyframe whose overlap with the latest keyframe is below a threshold $\tau_r$ or the frame whose camera distance from other keyframes is the farthest will be removed from the current keyframe window. By doing this, we update the Gaussians and networks with the knowledge from the new keyframes, which can be generalized better to a subsequent frame. 
\end{itemize}

\section{Losses} \label{sec:losses}
We describe in this section the detailed formulation of proposed regularizations applied on the avatar representation that are introduced in \secref{sec:method_slam}. 

\boldparagraph{Local Deformation Loss.} We constrain the local deformation to be as small as possible and encourage the frame-generic model to best learn the average shape and average. The local deformation loss $L_\text{deform}$ is composed of three parts that respectively penalize the displacement $\Delta \mu'$, push the rotation offset $\Delta R$ to be close to the identity matrix, and enforce the ambient occlusion factor $\Delta c$ to stay close to 1.
\begin{equation}
    L_\text{deform} = {\lVert \Delta \mu' \rVert}_2 + {\lVert \Delta R - R_{id} \rVert}_1 + {\lVert \Delta c  - 1\rVert}_2 \label{eq:loss_deform}
\end{equation}
In practice, we use quaternions to represent the rotations.

\boldparagraph{LBS Weights Loss.}
For each Gaussian in the canonical space, we use the K-Nearest Neighbor (KNN) algorithm to find the $k$ nearest SMPL vertices $v_\text{NN}$ and take the weighted sum of their LBS weights $\Tilde{W}$ as the label to supervise the Gaussian LBS weights with $L_\text{LBS} = {\lVert W - \Tilde{W} \rVert}_F$. Inspired by~\cite{hugs}, the displacements between the nearest k SMPL vertices and the Gaussian center, formulated as $\Delta v_{NN} = \mu + \Delta\mu - v_\text{NN}$, are used to weigh each element so that SMPL vertices closer to the corresponding Gaussian will contribute more to the supervision. Differently, we propose a novel distance-based weighting that takes account of the shape and scale of each Gaussian and calculate $\Tilde{W}$ as follows.
\begin{align}
    \Tilde{W} &= \sum_{i=1}^{k} \frac{w_i}{w} W_{\text{NN},i} \\
     w_i &= \exp{(-\frac{1}{2} {\Delta v_{\text{NN},i}}^T\Sigma^{-1} {\Delta v_{\text{NN},i}})}\\
     w &= \sum_{i=1}^{k} w_i
\end{align}

where $\Delta v_{\text{NN},i}$ and $W_{\text{NN},i}$ are respectively the relative position and LBS weight of the $i$-th nearest vertice on the SMPL mesh from the Gaussian center, and $\Sigma$ is the Gaussian covariance matrix that is defined as $\Sigma = RSS^{T}R^{T}$.

In our experiments, we set $k=3$.

\boldparagraph{Canonical Center Loss.} In our online pipeline, where the local window is typically small, each Gaussian is only visible to limited training views and is thus largely unconstrained. To prevent Gaussians from moving and growing arbitrarily along the camera ray, we softly regularize the geometry of the reconstructed human with the underlying SMPL model. Because the garment can lead to large displacements from naked-body SMPL to the reconstructed avatar, we do not directly regularize the magnitudes of the Gaussian displacements but instead enforce the nearest vertex on the SMPL mesh from each Gaussian to be the vertex used to initialize the Gaussian. The regularization is applied on the canonical Gaussian centers before local deformation as follows:
\begin{equation}
    L_\text{center} = \mathrm{ReLU}({\lVert\mu + \Delta\mu - v_\text{init}\rVert}_2 - {\lVert\mu + \Delta\mu - v_\text{NN}\rVert}_2)
\end{equation}

where $v_\text{NN}$ is the nearest SMPL vertex from the Gaussian in the canonical space, and $v_\text{init}$ is the corresponding vertex position initially.

We run the K-Nearest Neighbor algorithm via the efficient CUDA implementation in PyTorch3D~\cite{ravi2020pytorch3d}.

\section{Implementation Details}
\subsection{Model Configurations}
We initialize the canonical Gaussian positions by creating $r$ replicates of each SMPL vertex in the canonical space and injecting Gaussian noises. $r$ is set to be 5 for the EMDB dataset and 3 for the NeuMan dataset. The Gaussian opacities are initialized to be 0.9. We use the anisotropic Gaussians for both the scene and human parts.

For the LBS weights per Gaussian, we directly optimize an offset vector to sum to the original SMPL weights and use SoftMax as the activation function to apply to each element of the optimized weights to ensure that their values are all positive and sum to one.

For the time-pose dependent deformation network, the canonical points $\mu$ are first normalized with a bounding box that tightly encloses the canonical SMPL mesh. The detailed network hyperparameters are listed in \tabref{tab:hash_table}.

\begin{table}[h!]
    \centering
    \small
    \setlength\tabcolsep{26pt}
    \begin{tabular}{lr}
    \toprule
     Parameter & Value\\
    \midrule
    Number of levels & 16 \\
    Number of features per level & 2 \\
    Hash table size & $2^{17}$ \\
    Coarsest resolution & 4 \\
    Per Level Scale & 1.5 \\
    MLP Width & 128 \\
    MLP Number of hidden layers & 3 \\
    \bottomrule
    \end{tabular}
    \caption{Local deformation and ambient occlusion network hyperparameters}
    \label{tab:hash_table}
\end{table}

\subsection{Training Strategies}

\boldparagraph{Hash Encoding Network Pretraining.}
Random initial values of the hash encoding network can produce incorrect output on the fly when there are insufficient training frames and the training iterations are limited. This is the typical situation in the online training pipeline. Good interpolation and extrapolation properties are required to quickly fit the novel keyframe with the knowledge learned from previous frames. Otherwise, the Gaussian parameters could also get optimized in the wrong direction. 

Considering these issues, we propose to pre-train the local deformation and ambient occlusion networks introduced in \secref{sec:deform_nn} at the very beginning. This is achieved by randomly sampling input time and poses to obtain the deformation outputs from the hash encoding network and minimize the deformation loss $L_\text{deform}$. We sample the input time from a uniform distribution between 0 and 1. As for the human pose, we sample from a combination of the pose of the first frame and poses stored in a large-scale human database AMASS~\cite{AMASS:ICCV:2019} so that the network is pre-trained with realistic poses of large variations. Gaussian noises with a standard deviation of 0.1 are added to the input pose to augment the data. In our experiments, the poses in the BMLmovi dataset ~\cite{ghorbani2020movi} are used for sampling. We use Adam optimizer with learning rate $10^{-4}$ to run the optimization for 5000 iterations.

\boldparagraph{Multi-stage Training.}
We evenly divide the mapping process into two stages and choose not to include the time-pose-dependent deformation and ambient occlusion in avatar Gaussians in the first stage while later activate them in the second stage. This multi-stage training strategy is employed in both the online mapping and final color refinement steps.

\subsection{Training Configurations}
We use Adam Optimizer to optimize the camera and human pose parameters. The learning rates in the tracking thread are $3\times10^{-3}$ for camera rotation, $10^{-3}$ for camera translation, $10^{-2}$ for the human root translation and orientation, and $10^{-3}$ for other local pose parameters. In the mapping thread where we simultaneously perform local bundle adjustment on the keyframe window, the learning rates are reduced to $1.5\times10^{-3}$, $5\times10^{-4}$, $10^{-4}$ and $10^{-5}$ respectively. The learning rates of all the Gaussian parameters are exactly the same as the
original implementation from ~\cite{kerbl3Dgaussians}. For our additionally designed time-pose dependent network, we set learning rate of all its parameters to be $10^{-4}$.

In the tracking thread, we iteratively run camera and human pose optimization for 100 iterations with $\lambda_\text{rgb} = 1, \lambda_\text{flow} = 1, \lambda_\text{disp} = 0.001, \lambda_\text{sil} = 0.1, \lambda_\text{kp} = 0.0001$ in \eqref{eq:loss_pose}. While for mapping, we set $\lambda_\text{rgb} = 1, \lambda_\text{sil} = 1, \lambda_\text{depth} = 0.001, \lambda_\text{LBS} = 100, \lambda_\text{center} = 10, \lambda_\text{deform} = 0.001$ in \eqref{eq:loss_map}. Optimized Gaussians in the mapping thread are synchronized with the tracking thread every 20 mapping iterations. Finally when we iterate over the whole sequence, we finetune the Gaussians with all the selected keyframes for 100 epochs.

For keyframe selection, we set $\tau_t = 0.1$s, $\tau_c = 0.05$m, $\tau_j = 0.1$m, $\tau_v = 0.9$ as the thresholds. As for the local keyframe update, we set $\tau_s = 10$ and $\tau_r = 0.3$.

For the scene representation, we periodically perform Gaussian densification and pruning as originally described by 3DGS \cite{kerbl3Dgaussians}.  In contrast, for the fixed-size human, we disable the adaptive seeding during the online mapping since the complicated topology of the human body and the limited training viewpoints can lead to noisy gradients, especially in the occluded human parts. The densification and pruning module will be later activated for humans in the final color refinement step to capture richer details.

\subsection{Baselines}
When assessing the performance of novel view synthesis, we optimize human poses across all test frames for the baseline methods to eliminate the impact of pose errors on rendering. In contrast, for our approach, this step is omitted because the test poses are already optimized dynamically during the process. By adopting this strategy, we provide an advantage to the baselines, as their test poses are refined against the final reconstruction to minimize re-rendering errors. Conversely, our test poses are optimized using the online reconstructed model, which may be incomplete, sub-optimally refined, and therefore more susceptible to errors.

For consistency, we fix the Gaussian and network parameters across all methods and utilize each method's specific pose estimation module, applying the same loss functions used during their training to perform test pose optimization. This ensures that the evaluation of novel view synthesis reflects the robustness of the respective pose optimization designs as well. For direct pose estimation modules, as implemented in ~\cite{qian20233dgsavatar, xue2024hsr, guo2023vid2avatar, hugs}, we employ a uniform learning rate of $10^{-3}$. For the pose correction MLP network used in ~\cite{hu2023gauhuman}, we maintain the same learning rate as during training. To ensure fairness, pose optimization is conducted for 100 steps on each frame across all baselines.
\section{Additional Evaluation Results}
\subsection{Novel View Synthesis}

\begin{figure*}[!h]
    \centering
    
    \begin{subfigure}[b]{0.13\textwidth}
    \centering
        \includegraphics[width=\textwidth]{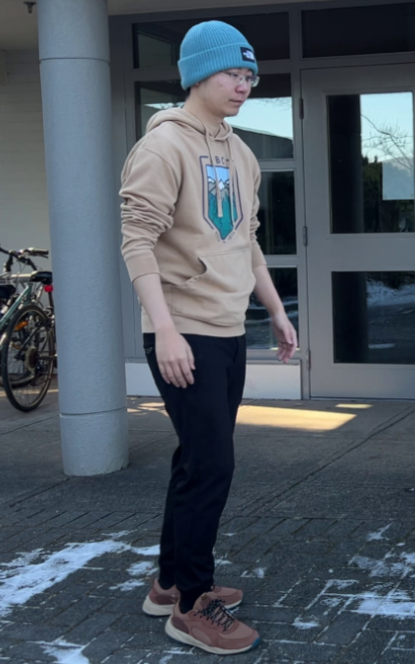} 
    \end{subfigure}
    \begin{subfigure}[b]{0.13\textwidth}
    \centering
        \includegraphics[width=\textwidth]{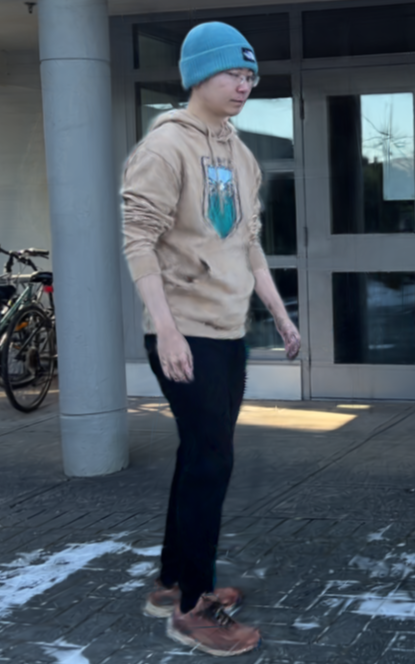} 
    \end{subfigure}
    \begin{subfigure}[b]{0.13\textwidth}
    \centering
        \includegraphics[width=\textwidth]{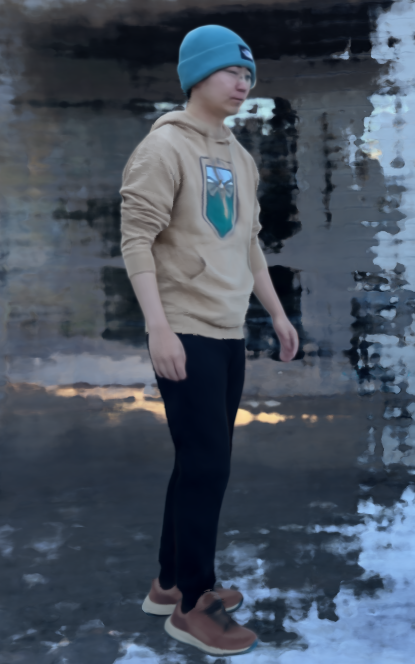}  
    \end{subfigure}
    \begin{subfigure}[b]{0.13\textwidth}
    \centering
        \includegraphics[width=\textwidth]{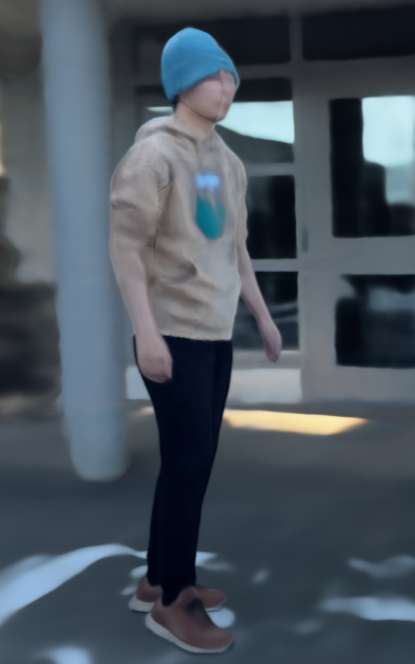} 
    \end{subfigure}
    \begin{subfigure}[b]{0.13\textwidth}
    \centering
        \includegraphics[width=\textwidth]{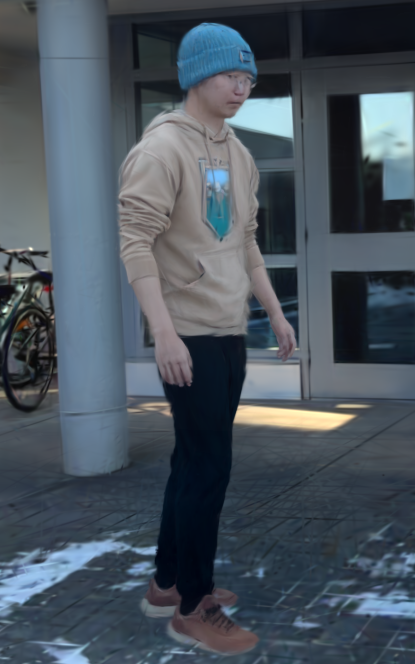} 
    \end{subfigure}
     \begin{subfigure}[b]{0.13\textwidth}
    \centering
        \includegraphics[width=\textwidth]{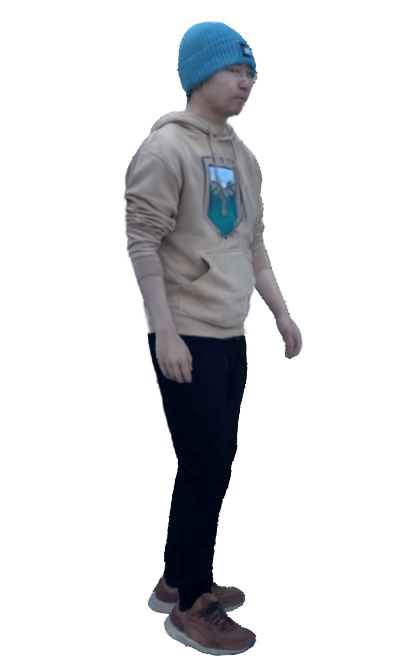} 
    \end{subfigure}
    \begin{subfigure}[b]{0.13\textwidth}
    \centering
        \includegraphics[width=\textwidth]{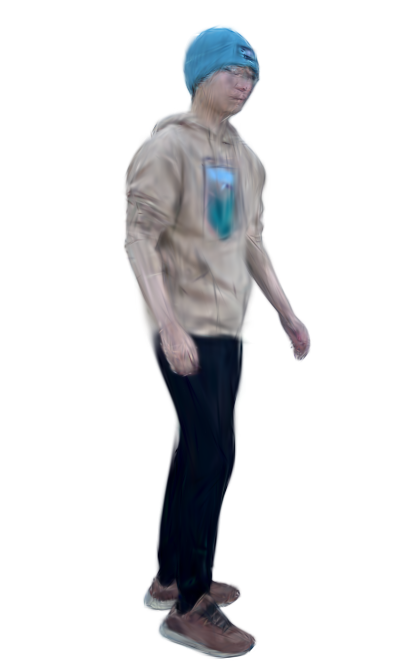} 
    \end{subfigure}
    \vskip\baselineskip
    \begin{subfigure}[b]{0.13\textwidth}
    \centering
        \includegraphics[width=\textwidth]{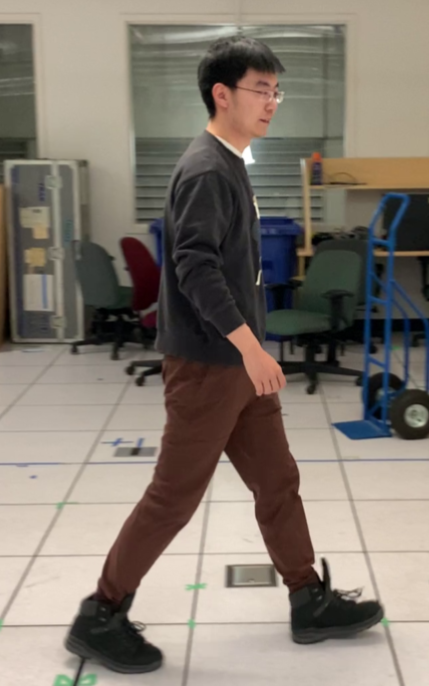} 
    \end{subfigure}
    \begin{subfigure}[b]{0.13\textwidth}
    \centering
        \includegraphics[width=\textwidth]{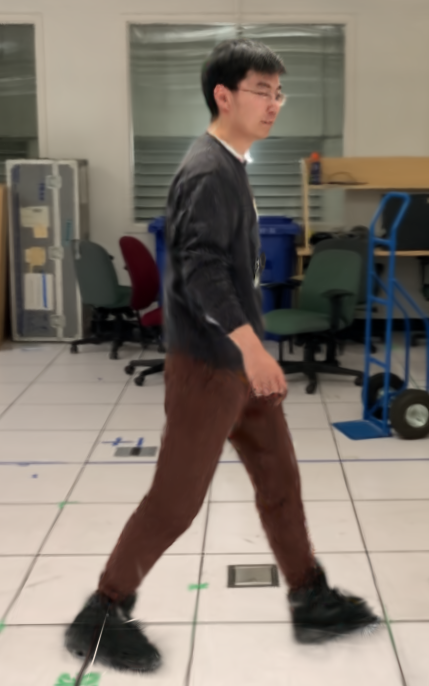} 
    \end{subfigure}
    \begin{subfigure}[b]{0.13\textwidth}
    \centering
        \includegraphics[width=\textwidth]{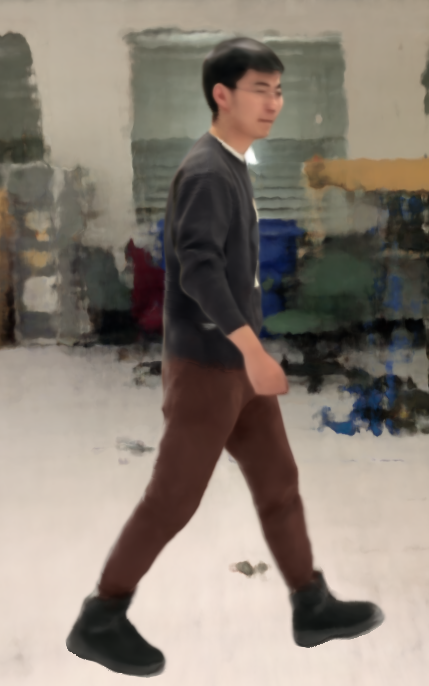}  
    \end{subfigure}
     \begin{subfigure}[b]{0.13\textwidth}
    \centering
        \includegraphics[width=\textwidth]{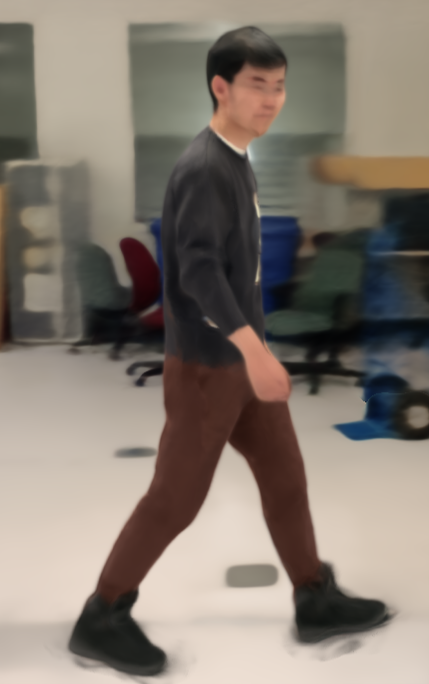} 
    \end{subfigure}
    \begin{subfigure}[b]{0.13\textwidth}
    \centering
        \includegraphics[width=\textwidth]{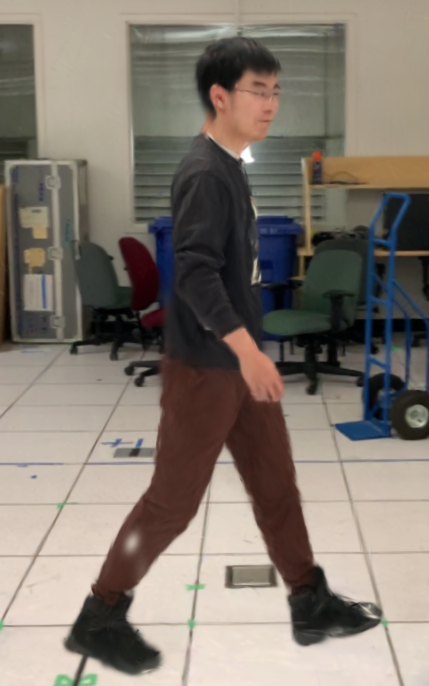} 
    \end{subfigure}
     \begin{subfigure}[b]{0.13\textwidth}
    \centering
        \includegraphics[width=\textwidth]{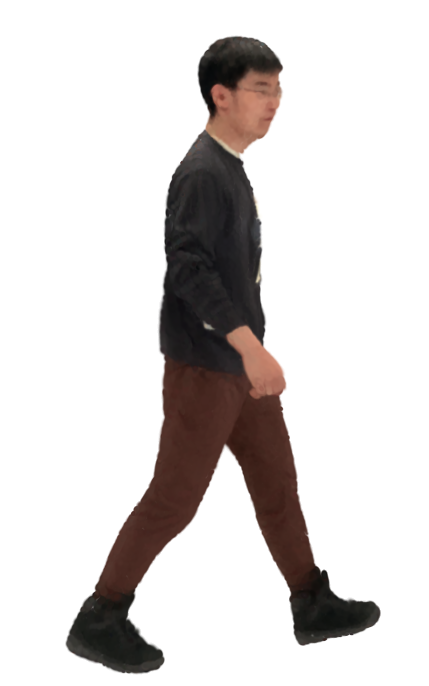} 
    \end{subfigure}
    \begin{subfigure}[b]{0.13\textwidth}
    \centering
        \includegraphics[width=\textwidth]{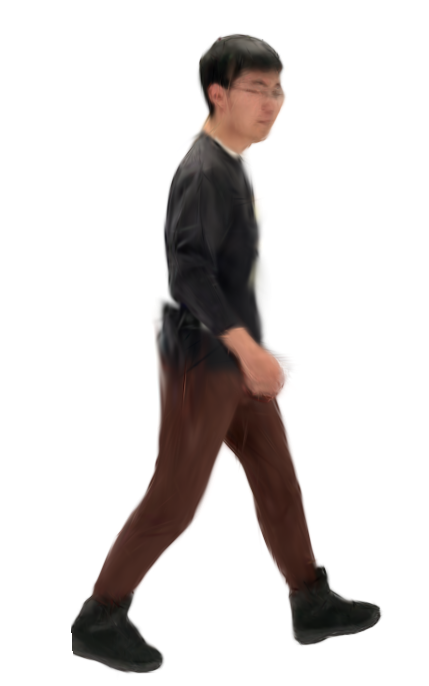} 
    \end{subfigure}
    
    \vskip\baselineskip
    \begin{subfigure}[b]{0.13\textwidth}
    \centering
        \includegraphics[width=\textwidth]{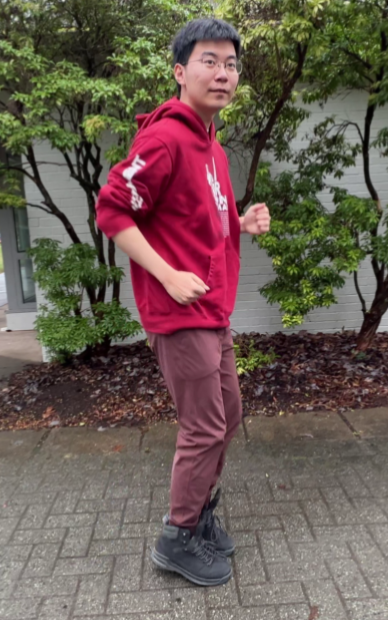}
        \caption[]{GT}    
    \end{subfigure}
    \begin{subfigure}[b]{0.13\textwidth}
    \centering
        \includegraphics[width=\textwidth]{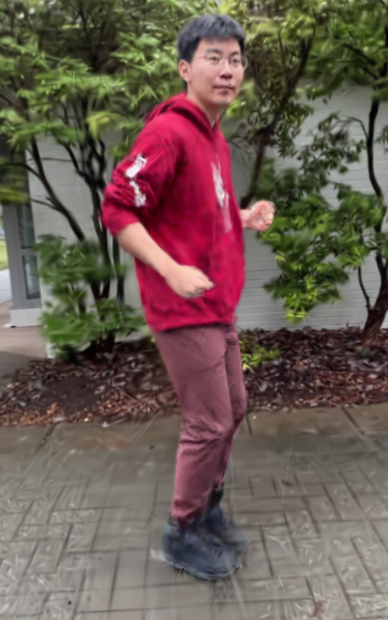}
        \caption[]{Ours}    
    \end{subfigure}
    \begin{subfigure}[b]{0.13\textwidth}
    \centering
        \includegraphics[width=\textwidth]{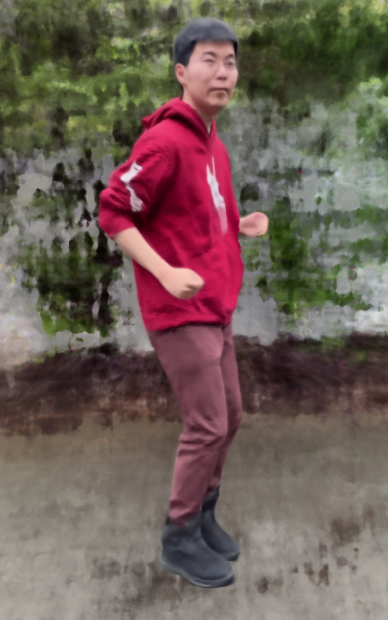}
        \caption[]{Vid2Avatar}    
    \end{subfigure}
    \begin{subfigure}[b]{0.13\textwidth}
    \centering
        \includegraphics[width=\textwidth]{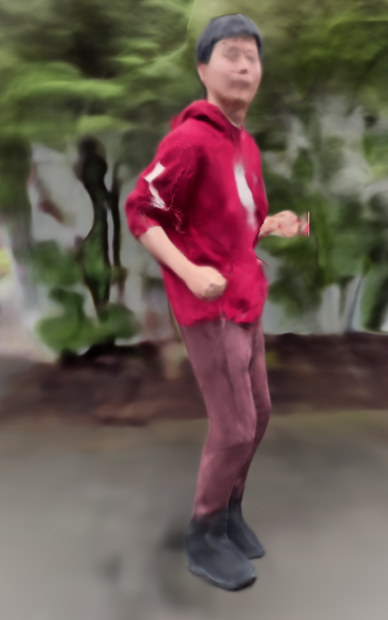}
        \caption[]{HSR}    
    \end{subfigure}
    \begin{subfigure}[b]{0.13\textwidth}
    \centering
        \includegraphics[width=\textwidth]{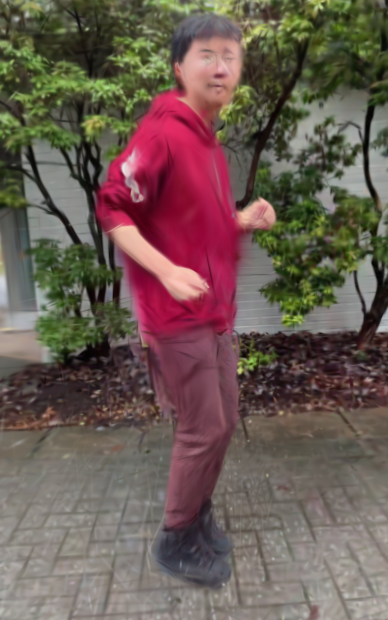}
        \caption[]{HUGS}    
    \end{subfigure}
     \begin{subfigure}[b]{0.13\textwidth}
    \centering
        \includegraphics[width=\textwidth]{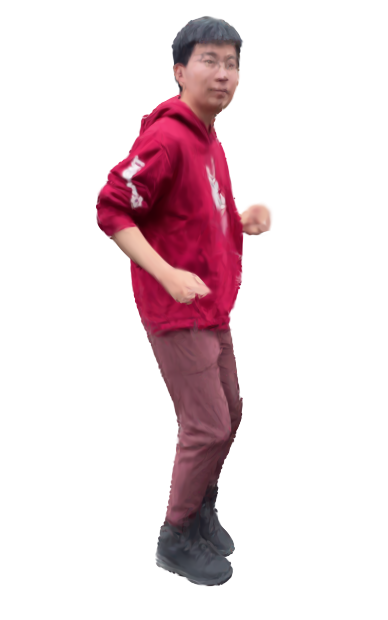}
        \caption[]{3DGS-Avatar}    
    \end{subfigure}
    \begin{subfigure}[b]{0.13\textwidth}
    \centering
        \includegraphics[width=\textwidth]{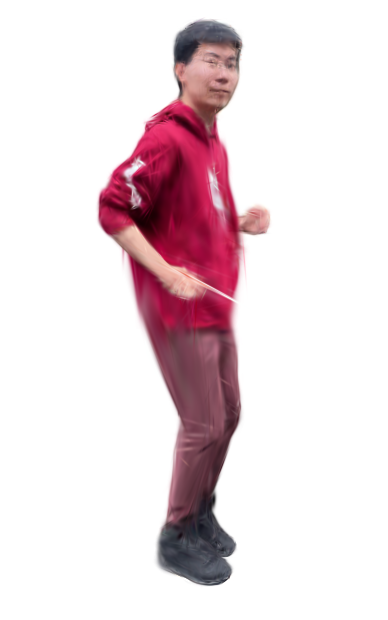}
        \caption[]{GauHuman}    
    \end{subfigure}

    \caption{Qualitative comparison of novel view synthesis task on the NeuMan dataset~\cite{jiang2022neuman}.}
    \label{fig:neuman_novel_view}
\end{figure*}

Qualitative results on the NeuMan dataset~\cite{jiang2022neuman} are presented in \figref{fig:neuman_novel_view}. Despite performing online tracking and mapping, our method surpasses most offline reconstruction approaches in terms of background scene fidelity and clarity, even though those methods leverage ground truth camera poses. Furthermore, our approach achieves superior quality in the reconstruction of critical and challenging human features, such as faces and hands. However, a limitation of our method is that geometry near contact points between the human and the scene may not always be precisely recovered, occasionally resulting in blurry reconstructions, as seen in areas like shoes and the ground.

\subsection{Camera Tracking}

\begin{table}[t]
    \centering
    \footnotesize
    \begin{tabular}{lc}
    \toprule
    & ATE RMSE [m]$\downarrow$ \\  \midrule
    DROID-SLAM & \fs 0.079 \\
	MonoGS (human masked) & 0.459 \\
    Ours (human masked) & \rd 0.247 \\
    Ours (full model) & \nd 0.084 \\

    \bottomrule
    \end{tabular}
    \caption{Camera tracking evaluation on the EMDB dataset.}
    \label{tab:emdb_camera}
\end{table}

We demonstrate that our camera tracker achieves on-par performance with the state-of-the-art SLAM approach DROID-SLAM\cite{teed2022droidslamdeepvisualslam} in \tabref{tab:emdb_camera}. Without knowing the true scale, the output from DROID-SLAM cannot be seamlessly integrated with human pose estimates unless ground truth depth or trajectory information is provided, limiting its applicability in dynamic scenes. However, by explicitly building the dynamic human and modeling human-scene spatial correlation, our method handles the scaling well. Moreover, to further inspect the impact of human on the tracking, we run MonoGS\cite{Matsuki2024MonoGS} and our method while using pre-estimated human masks to completely remove the human in the input images and the model. As shown in \tabref{tab:emdb_camera}, our method significantly enhances the accuracy of predicted camera trajectories by explicitly modeling the human, as it provides additional spatial cues and aids in scaling the monocular depth signal.

\subsection{Human Pose Estimation}
\begin{figure}[!h]
    \centering
     \includegraphics[width=0.45\textwidth]{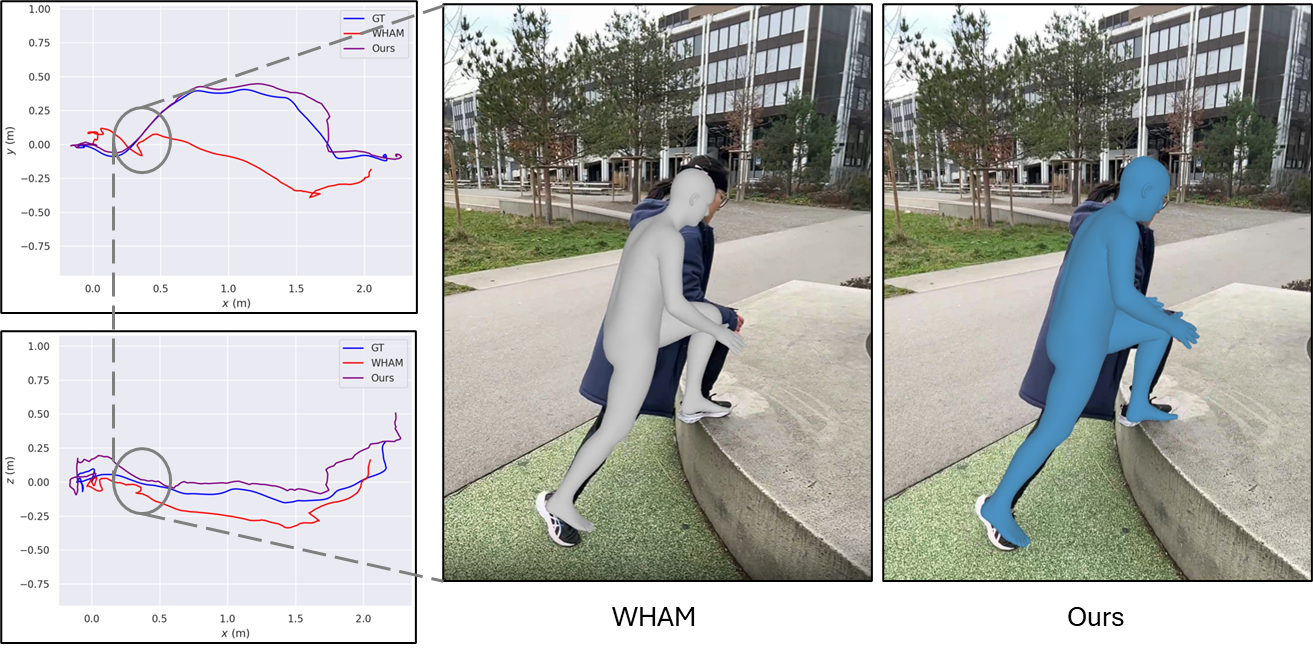}  
     \caption{Comparison of global human trajectory estimations on the EMDB dataset. Left: Human trajectories of \textcolor{blue}{GT}, \textcolor{red}{WHAM} predictions and \textcolor{purple}{our} predictions on the x-y and x-z plane. The global trajectories are globally aligned. Right: Estimated SMPL mesh on one selected frame.} \label{fig:human_traj}
\end{figure}

We evaluate our human pose estimations and compare them with WHAM\cite{shin2023wham} in \tabref{tab:emdb_pose} and \figref{fig:human_traj}. Our reconstruction-based pose optimization module achieves slightly enhanced local poses that align more accurately with the 2D image. For global motion, our holistic human-scene reconstruction supplies the essential spatial context, enabling the human tracker to significantly reduce globally aligned joint errors. In contrast, WHAM, lacking explicit scene awareness, fails to adapt to terrain changes, resulting in substantial trajectory errors. However, the increased jitter observed in our method indicates a limitation: the gradient descent optimization approach becomes ineffective for occluded body parts that are not visible in the 2D image.

\begin{table*}[t]
    \centering
    \footnotesize
    \begin{tabularx}{\textwidth}{lCCCCCC}
    \toprule
    & \multicolumn{3}{c}{\textbf{Local Pose}} & \multicolumn{3}{c}{\textbf{Global Motion}} \\
    & PA-MPJPE$\downarrow$ & MPJPE$\downarrow$ & MVE$\downarrow$ & Jitter$\downarrow$ & WA-MPJPE$\downarrow$ & W-MPJPE$\downarrow$ \\
    \midrule
    WHAM & 40.845 & 72.964 & 83.254 & \textbf{14.765} & 636.001& 2990.746\\
    Ours & \textbf{40.571} & \textbf{69.162} & \textbf{79.463} & 32.183& \textbf{175.215}& \textbf{449.036}\\

    \bottomrule
    \end{tabularx}
    \caption{Human pose estimation evaluation on the EMDB dataset. Jitter is in the unit of 10$m/s^{-3}$ and others in $mm$.}
    \label{tab:emdb_pose}
\end{table*}

\section{Ablation Study}
\subsection{Ablation of Avatar Module Designs}
Input and output components of the avatar deformation module are ablated in \tabref{tab:abl_avatar}. On the challenging EMDB dataset where drastic garment deformation and illumination change exist, jointly modeling the per-Gaussian deformation and ambient occlusion significantly improves all the re-rendering metrics. As for the input, we achieve the best performance by taking both the pose and time features compared to using either one of them. 

\begin{table}[h!]
    \centering
    \begin{tabular}{lccc}
    \toprule
    & PSNR $\uparrow$ & SSIM $\uparrow$ & LPIPS $\downarrow$ \\
    \midrule
    w/o ambient occlusion & \rd 28.201 & 0.958 & \rd 0.034\\
    w/o deformation & 27.927 &  \rd 0.959 & 0.036 \\
    w/o pose encoding & \nd 28.741 &  \nd 0.962 & \nd 0.033 \\
    w/o time encoding & 27.779 & 0.957 & 0.037\\ 
    w/o HE pretraining & 27.801 & 0.958 & 0.041 \\ 
   Full model & \fs 28.955  &  \fs 0.966 & \fs 0.031 \\

    \bottomrule
    \end{tabular}
    \caption{Ablation study on avatar module designs and hash encoding (HE) network pretraining strategy. The performance is evaluated on the \textbf{human-only} rendering on the EMDB dataset.}
    \label{tab:abl_avatar}
\end{table}

\subsection{Ablation of Hash Encoding Network Pretraining Strategy}

In \tabref{tab:abl_avatar}, we also present the evaluation results without pretraining the hash encoding network. Due to the randomized initial network parameters, the local deformation network produces noisy outputs, resulting in failed learning of garment deformation and shadows, particularly at unseen timesteps and poses. The bad interpolation and extrapolation properties lead to an overall degraded performance.

\section{Discussions}

\subsection{Online Training}
We follow existing dense SLAM works\cite{Matsuki2024MonoGS, teed2022droidslamdeepvisualslam, keetha2024splatam} to perform a final refinement step to finetune the Gaussian representation with all the selected keyframes. The refinement process can be seen as a traditional global bundle adjustment (BA) step, in which case it does not conflict with the online nature of \methodname. Unlike other approaches, we do not perform full BA but instead refine only the Gaussians, allowing us to distribute the refinement into the online optimization rather than applying it as a post-processing step—though this comes at the cost of lower training FPS. By distributing refinement into the online pipeline after each keyframe tracking step and training for ten epochs per refinement operation, we achieve a final PSNR of 23.013 for the whole image and 28.814 for human-only regions, which is slightly worse than the full model and increases runtime (reducing FPS by 0.06). The final refinement step is designed to prevent catastrophic forgetting, and we showcase that without this, \methodname still largely overperforms baselines in novel view synthesis and runtime efficiency.

\subsection{Challenging Cases}

\begin{figure}[!h]
    \centering
     \includegraphics[width=0.45\textwidth]{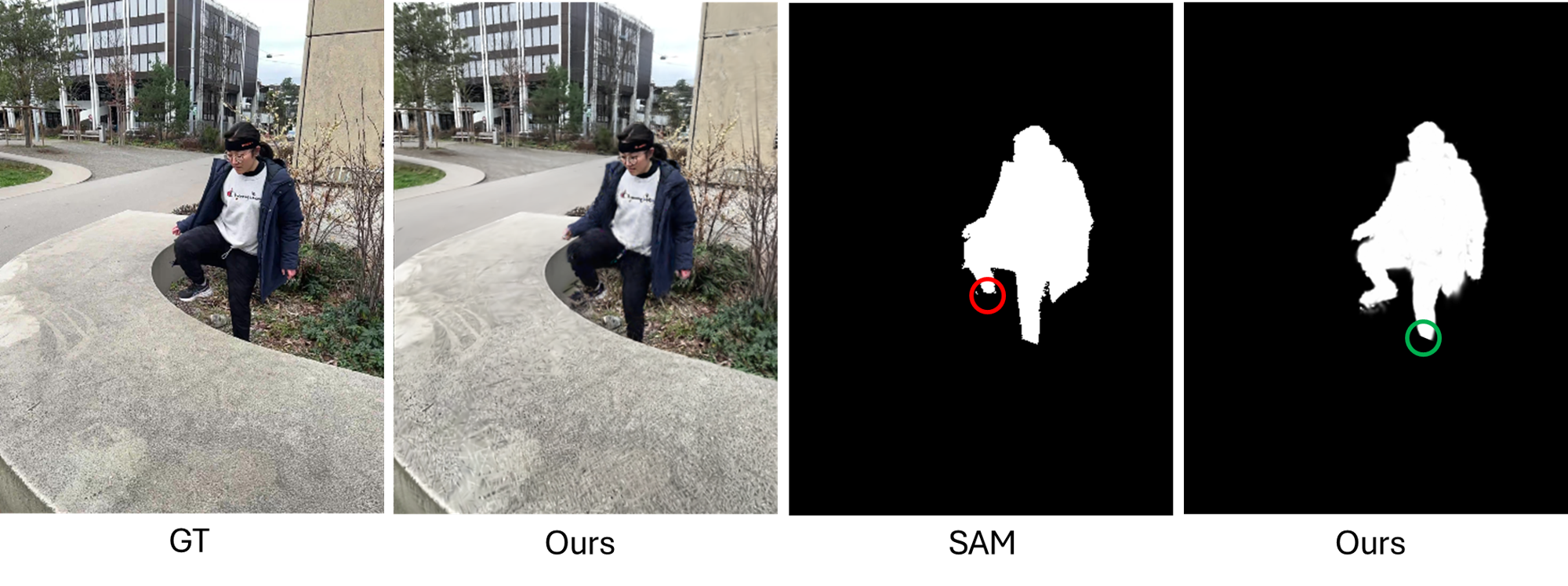}  
     \caption{Results in the scene occlusion scenario. Our generated human mask is compared against the prediction from the Segment Anything Model(SAM).} \label{fig:scene_occlusion}
\end{figure}

\boldparagraph{Scene Occlusion.} We demonstrate the impact of our occlusion-aware human silhouette design in \figref{fig:scene_occlusion}. For body parts occluded by scene components, such as legs, \methodname consistently generates smooth and precise boundary silhouettes. In contrast, the state-of-the-art general segmentation model SAM\cite{ren2024groundedsamassemblingopenworld}, while capable of predicting occlusions, occasionally produces results with missing parts. By explicitly modeling occlusions, \methodname effectively models spatial correlations without losing human features.

\begin{figure*}[!h]
    \centering
     \includegraphics[width=\textwidth]{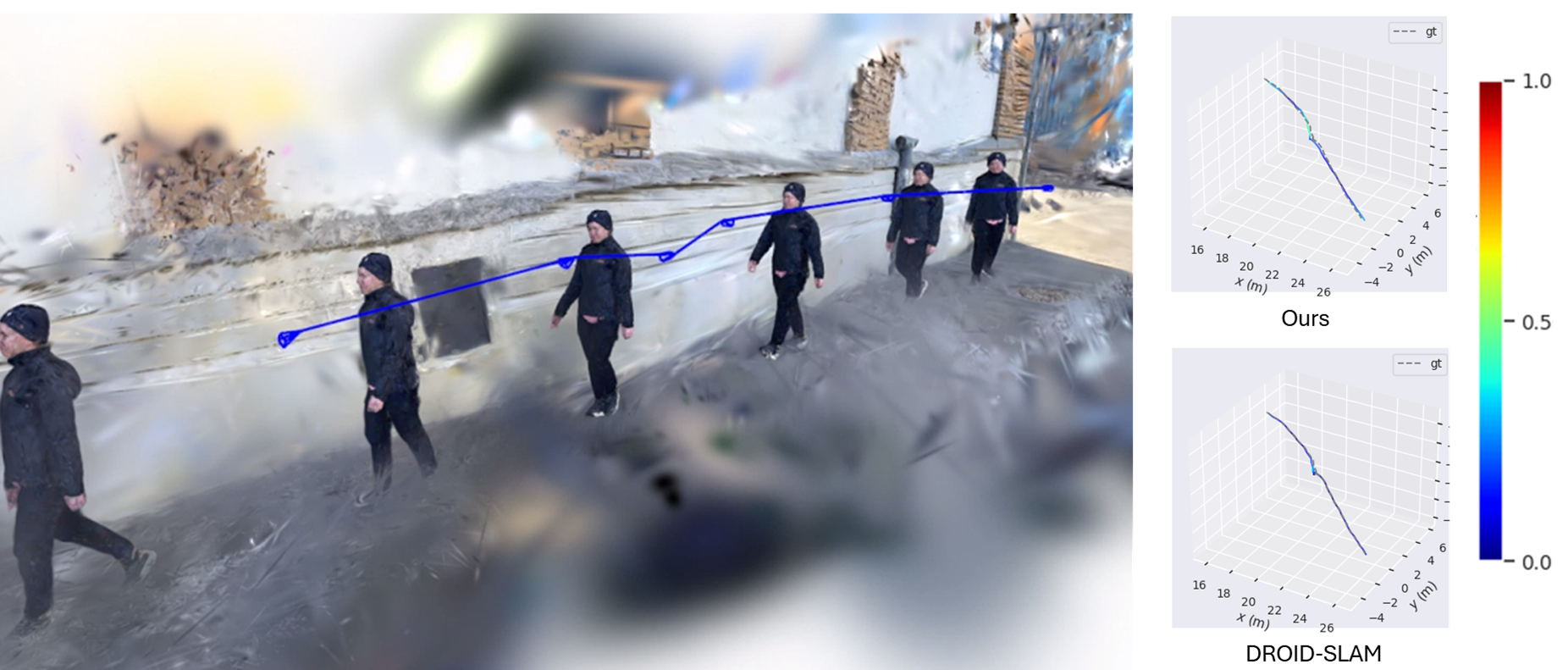}  
     \caption{Results in the long trajectory scenario. Left: Our human-scene reconstruction with tracked cameras. Right: Estimated trajectories from ours and DROID-SLAM, compared with the ground truth on the EMDB dataset. Colors of the curve segments indicate trajectory error, ranging from \textcolor{blue}{0.} to \textcolor{red}{1.}} \label{fig:long_traj}
\end{figure*}

\boldparagraph{Long Trajectories.} In \figref{fig:long_traj}, we showcase the results of our method in a long-trajectory scenario, where repetitive background patterns pose challenges for camera tracking. Overall, \methodname delivers decent results and effectively captures camera motion trends with small trajectory errors. However, the sparse features on the wall and ground increase the challenge of accurate geometric reconstruction, introducing some surface noise that subsequently leads to additional errors in the estimated camera poses for certain frames. DROID-SLAM performs better in such scenarios by leveraging cleverer bundle adjustment and graph-based optimization strategy, highlighting a promising direction for further improvements.

\subsection{Limitations}
While \methodname achieves state-of-the-art rendering quality on the challenging in-the-wild dataset, its performance heavily depends on single-frame pre-estimations, such as monocular depth and human keypoints—particularly in the first frame, which initializes the system. Although we incorporate a pairwise flow loss in camera and human pose optimization, we argue that this alone is insufficient for constructing a globally consistent scene and pose representation. Also, despite producing high-quality renderings, our method introduces surface noise due to the nonsmooth depth characteristics of 3D Gaussian Splatting. Additionally, our method could suffer from potential human-scene interpenetrations around the contact points, such as feet. Due to the noisy surfaces 3D Gaussians produce, it is not yet resolved. Finally, our model-based camera and human pose optimization primarily relies on pixel-level errors, which can lead to local optima in textureless regions or areas with uniform features, such as walls and clothing.

\end{document}